\documentclass[10pt,twocolumn,letterpaper]{article}

\usepackage[pagenumbers]{cvpr} % To force page numbers, e.g. for an arXiv version
\usepackage{bbding} % added by jiahui
\usepackage{multirow} % added by jiahui
\definecolor{cvprblue}{rgb}{0.21,0.49,0.74}
\usepackage[pagebackref,breaklinks,colorlinks,allcolors=cvprblue]{hyperref}

\def\confName{CVPR}
\def\confYear{2026}

\title{Towards Source-Aware Object Swapping with Initial Noise Perturbation}
\author{Jiahui Zhan, Xianbing Sun\textsuperscript{*}, Xiangnan Zhu,  Yikun Ji, Ruitong Liu, Liqing Zhang, Jianfu Zhang\textsuperscript{$\dagger$}\\
Shanghai Jiao Tong University
}

\begin{document}
% \maketitle

\twocolumn[{ 
\renewcommand\twocolumn[1][]{#1}%
\maketitle  
\begin{center}
    \centering
    \captionsetup{type=figure}
    \includegraphics[width=0.99\linewidth]{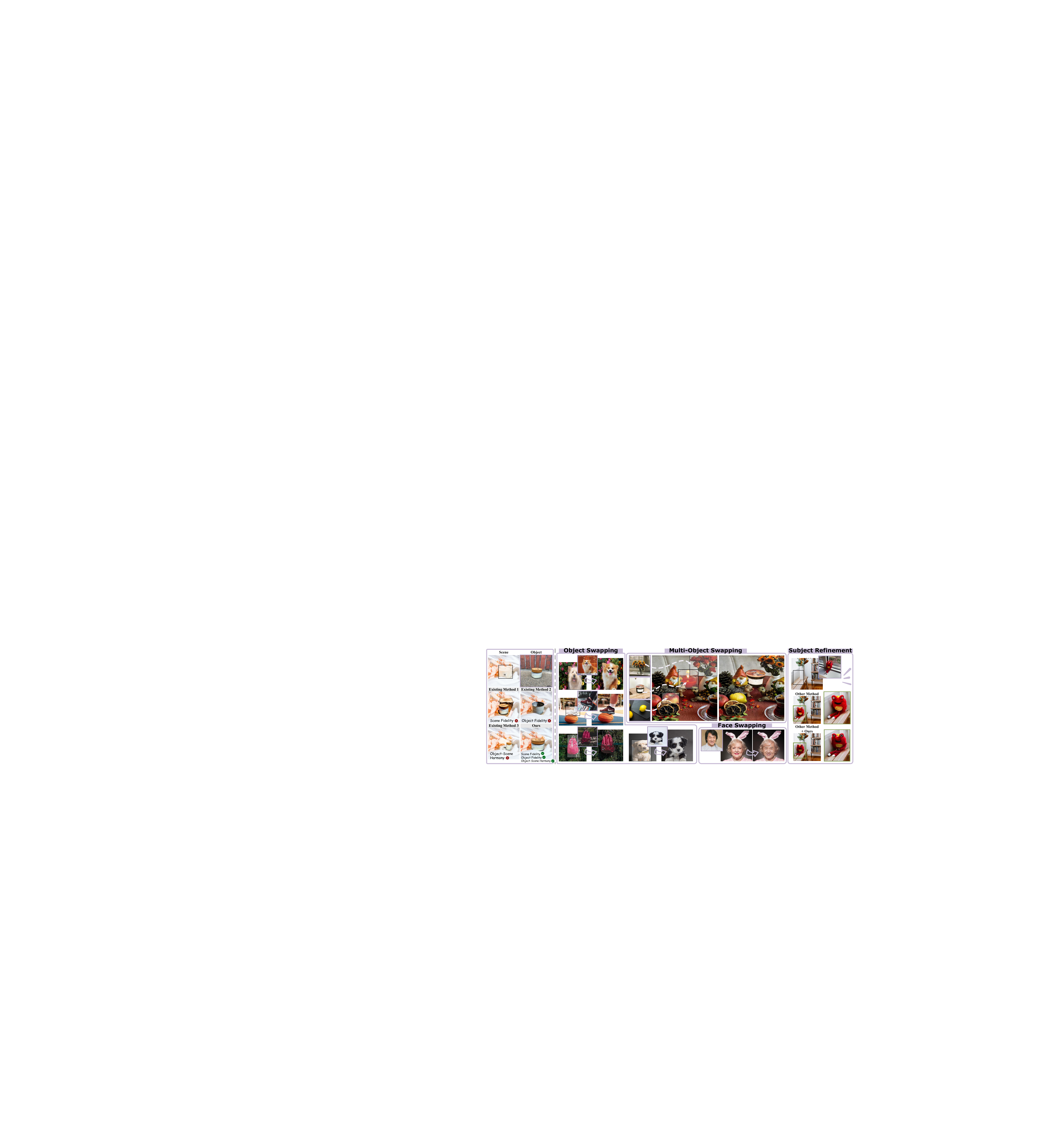}
    \captionof{figure}{
    We propose \emph{SourceSwap}, a source-aware object swapping framework that learns inter-object alignment from single images, achieving superior object fidelity, scene preservation, and object–scene harmony over prior methods. It is versatile in use, supporting multi-object swapping, face swapping, and subject-driven refinement.
    }\label{fig:teaser}
\end{center}
% \vspace{1em}
}]

\begingroup
  \renewcommand\thefootnote{}
  \footnotetext{\textsuperscript{*}Equal contribution; \textsuperscript{$\dagger$}Corresponding author (c.sis@sjtu.edu.cn).}
\endgroup

\begin{abstract}
Object swapping aims to replace a source object in a scene with a reference object while preserving object fidelity, scene fidelity, and object–scene harmony. 
Existing methods either require per-object finetuning and slow inference or rely on extra paired data that mostly depict the same object across contexts, forcing models to rely on background cues rather than learning cross-object alignment.
We propose \textit{SourceSwap}, a self-supervised and source-aware framework that learns cross-object alignment.
Our key insight is to synthesize high-quality pseudo pairs from \emph{any} image via a frequency-separated perturbation in the initial-noise space, which alters appearance while preserving pose, coarse shape, and scene layout, requiring no videos, multi-view data, or additional images.
We then train a dual U-Net with full-source conditioning and a noise-free reference encoder, enabling direct inter-object alignment, zero-shot inference without per-object finetuning, and lightweight iterative refinement.
We further introduce \textit{SourceBench}, a high-quality benchmark with higher resolution, more categories, and richer interactions. Experiments demonstrate that \textit{SourceSwap} achieves superior fidelity, stronger scene preservation, and more natural harmony, and it transfers well to edits such as subject-driven refinement and face swapping.

\end{abstract}    

\section{Introduction}
\label{sec:intro}
The surge of diffusion models~\cite{sd,podellsdxl,wu2025qwen} has naturally pushed controllable generation~\cite{controlnet,t2iadapter} to the forefront of visual synthesis.
As a representative task in controllable generation, subject-driven generation~\cite{dreambooth,TextualInversin} synthesizes images conditioned on specific subjects rather than relying solely on coarse textual prompts.
Such explicit control opens up broad opportunities for content creation and visual design, with wide applications in e-commerce, advertising, and digital media.
We study a more challenging setting, \emph{object swapping}~\cite{lidreamedit,gu2023photoswap,gu2024swapanything,zhuinstantswap}. Given a source image, the goal is to replace the source object with a reference object from another image while keeping the scene realistic and consistent. A successful swap should satisfy three criteria: \textit{object fidelity} (the swapped object preserves the identity details of the reference), \textit{scene fidelity} (the background remains unchanged and boundary regions are minimally perturbed), and \textit{object–scene harmony} (the swapped object adapts to the illumination, viewpoint, and style of the scene and matches the morphology of the original source object). \looseness=-1

Existing solutions fall short when measured against these goals. 
Test-time tuning methods~\cite{lidreamedit,gu2023photoswap,zhuinstantswap,gu2024swapanything} extend personalization~\cite{dreambooth,TextualInversin} to fixed scenes but require per-object finetuning and are sensitive to hyperparameters. Over- or under-fitting harms object fidelity, the tuned concept often fails to align with the source morphology and local interactions (harming harmony), and background preservation relies on accurate inversions, leading to weak scene fidelity in practical settings.
Training-free editing methods~\cite{TIGIC,lu2023tf-icon,diptychprompt} rely on prompt tricks or pipeline tweaks and lack global reasoning over multiple inputs (source, reference, mask), so they frequently miss fine identity cues and produce seams or lighting mismatches around the edited region.
Learning‑based inpainting methods \cite{pbe,song2023objectstitch,chen2024anydoor,mimicbrush,song2024imprint,winter2025objectmate} improve scene fidelity with pretrained inpainting models but depend on pseudo pairs built from cropping \cite{pbe,chen2024anydoor,song2023objectstitch}, video frames \cite{chen2024anydoor,mimicbrush}, retrieval \cite{winter2025objectmate}, or multi‑view data \cite{song2024imprint}. 
Video pairs bring blur and low resolution, harming object fidelity, while crop/retrieval/multi-view pairs mostly show the same object across contexts rather than relations between distinct objects, so the model does not learn the morphology matching needed for object–scene harmony.
Moreover, most methods mask the source object during training and inference, forcing the model to infer object state from background cues. This weakens cross-object alignment and introduces unnatural boundaries, reducing both harmony and scene fidelity. 
\looseness=-1

To address these challenges, we propose \textit{SourceSwap}, a self-supervised and source-aware object swapping framework that explicitly learns inter-object alignment.
First, to overcome the lack of suitable training pairs, SourceSwap introduces a \emph{frequency-separated perturbation in the initial-noise space} that turns \emph{any} image into a high-quality pseudo pair, changing appearance while preserving coarse shape and scene context, without relying on videos, multi-view data, or extra images.
This avoids the blur and misalignment issues of video-based pairs and, unlike crop/retrieval/multi-view schemes, exposes relations between distinct objects instead of only the same object across contexts.
Then, SourceSwap trains a dual U-Net with full-source conditioning and a clean reference encoder, removing the explicit source mask. The denoiser observes the whole source image rather than only a masked hole, which strengthens cross-object alignment, maintains scene fidelity, and improves object–scene harmony.
Full-source conditioning also naturally enables \emph{iterative refinement}: by feeding the previous output as the next input for a few rounds, we further improve identity fidelity while preserving scene details, with fast, zero-shot inference and no per-object finetuning.
Beyond our proposed benchmark \textit{SourceBench}, SourceSwap also delivers strong results on the widely used DreamEditBench~\cite{lidreamedit}, demonstrating robust performance across datasets with different object categories, resolutions, and interaction patterns.
It further transfers well to edits such as multi-object swapping, subject-driven refinement, and face swapping (see \cref{fig:teaser}).

Overall, our contributions are summarized as follows:
\begin{itemize}

    \item Single-image pairing via frequency-separated initial noise perturbation: any image to a high-quality pair, coarse shape and scene context preserved, enabling learning inter-object alignment directly.
    \item Dual U-Net with a noise-free reference encoder and full source conditioning enabling \textit{SourceSwap} zero-shot inference without per-object finetuning, fast and amenable to light iterative refinement.
    \item \emph{SourceBench}, a novel benchmark with higher resolution, more categories, and richer object–scene interactions. \looseness=-1
    \item Extensive experiments show \textit{SourceSwap}'s strong fidelity, scene preservation, and harmony, with seamless transfer to subject-driven refinement and face swapping.
	
\end{itemize}
\section{Related Works}
\begin{figure}
	\centering
	\includegraphics[width=\linewidth]{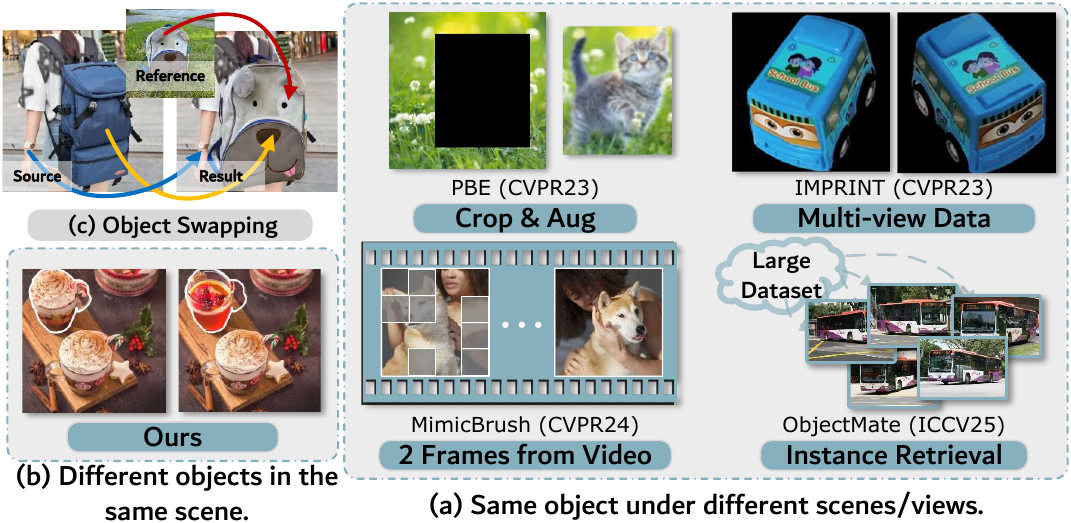}
	\caption{
Motivation of initial-noise perturbation. 
Existing inpainting-based methods rely on paired data from multi-view capture, videos, retrieval, or cropping, 
which are costly, blurry, or cannot model meaningful appearance changes of distinct objects. 
Our approach generates high-quality pseudo pairs from single images, 
producing clear yet coherent object variations while keeping the background intact.
    }
	\label{fig:moti}
    \vspace{-0.3cm}
\end{figure}

\begin{figure*}
	\centering
	\includegraphics[width=\linewidth]{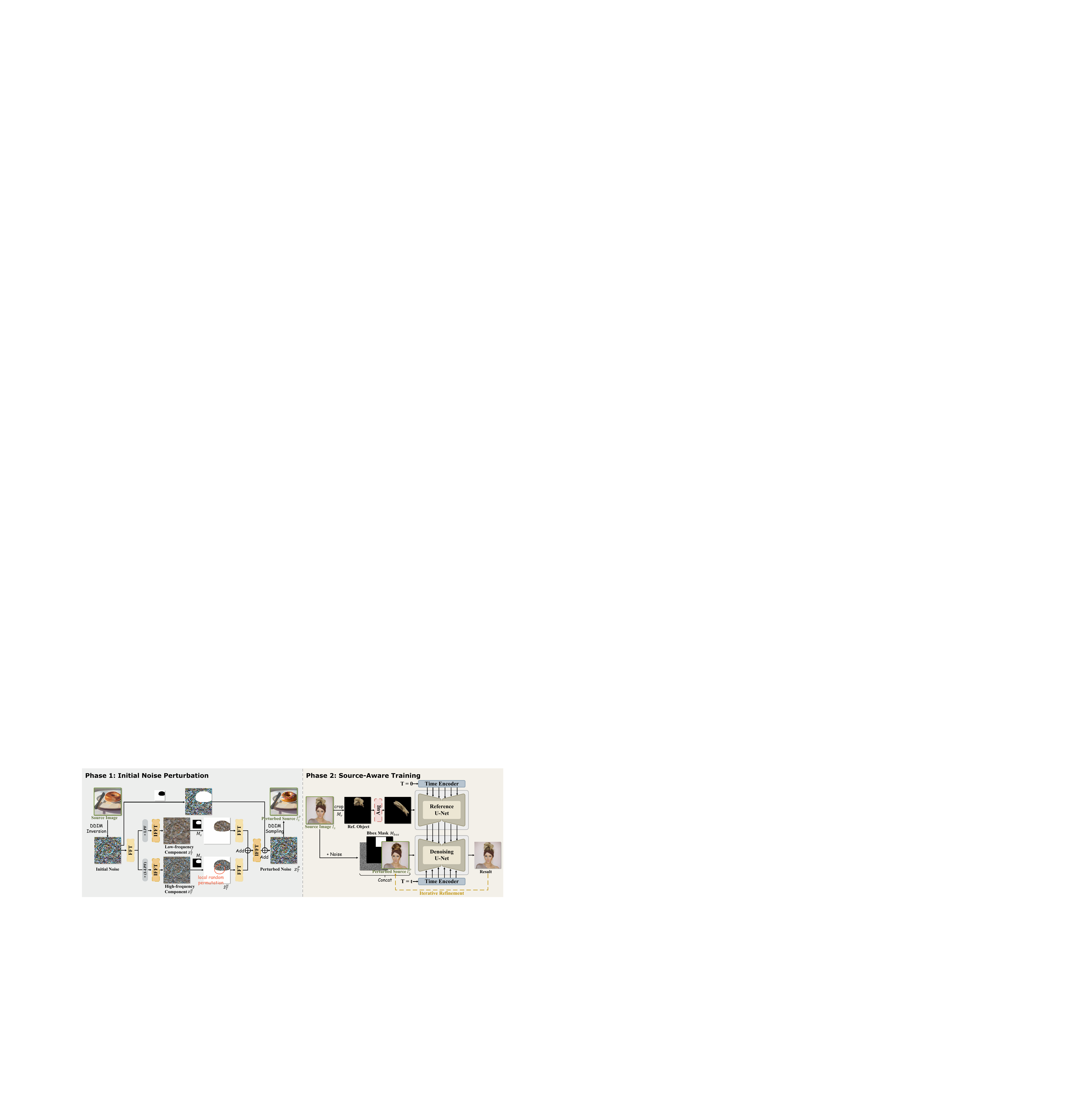}
    % \vspace{-0.4cm}
	\caption{
    Overview of \textit{SourceSwap}. The pipeline consists of two phases: 
    (1) Initial-noise perturbation, which converts any source image $I_s$ into a pseudo pair by transforming its initial latent $z_T$ to the frequency domain and locally permuting high-frequency components within the object mask to alter appearance while preserving coarse structure and scene context;
    and (2) Source-aware training, which uses these pseudo pairs to train a dual U-Net with full-source conditioning and a clean-reference encoder, enabling efficient learning of inter-object alignment and zero-shot inference. A lightweight iterative refinement step further improves object fidelity.
    }
	\label{fig:pipeline}
    \vspace{-0.4cm}
\end{figure*}

\noindent\textbf{Personalized Image Generation.}
For object swapping, a direct route is personalization \cite{dreambooth,TextualInversin,CD}: encode the reference object as a ``concept'' in a text-to-image model and then render it, but this does not natively compose the subject into a user-specified background. 
To bridge this gap, recent works combine personalized models with inversion to keep the background fixed while injecting the subject.
DreamEdit~\cite{lidreamedit} proposes iterative update pipeline to incorporate the concept, 
Photoswap~\cite{gu2023photoswap} replaces attention maps to inject the target subject into the background, 
SwapAnything~\cite{gu2024swapanything} further disentangles and fuses object and background latent variables to achieve seamless replacement, 
InstantSwap~\cite{zhuinstantswap} refines the score distillation to handle sharp shape disparity.
However, they inherit the core drawbacks of test‑time personalization: few‑shot data requirements, slow and hyperparameter‑sensitive inference, and heavy dependence on the quality of the pretrained model. 
In contrast, our \textit{SourceSwap} performs fast and stable zero-shot inference without any object-specific tuning.

\noindent\textbf{Subject-driven Image Inpainting.}
Another practical path to object swapping is subject‑driven inpainting: 
given a source image and a mask, fill the region with the reference object while keeping nearby pixels.
Training-free variants modify the sampling pipeline or prompts~\cite{TIGIC,lu2023tf-icon,diptychprompt} but lack joint reasoning over multiple inputs and often produce low-quality edits. 
Learning-based methods train dedicated inpainting models, typically with dual streams for the reference and the masked background (PBE~\cite{pbe}, ObjectStitch~\cite{song2023objectstitch}, AnyDoor~\cite{chen2024anydoor}). 
Because paired supervision is missing, they synthesize pseudo pairs via foreground–background crops (PBE, ObjectStitch), adjacent video frames (AnyDoor, MimicBrush~\cite{mimicbrush}), multi-view images (IMPRINT~\cite{song2024imprint}), or instance retrieval (ObjectMate~\cite{winter2025objectmate}). 
These choices limit transformations, introduce blur, or lack complete backgrounds, and they still require explicit masks at train and test time. As a result, models tend to infer object state from background cues rather than learn cross-object alignment, leading to boundary artifacts and weak harmony. 
In contrast, \textit{SourceSwap} removes the explicit source mask during training and learns direct inter-object alignment from robust single-image pseudo pairs, improving boundary quality and scene consistency.

\noindent\textbf{Guidance In Initial Noise Space.}
Prior studies \cite{guo2024initno,NoiseRefine,NPNet,NoiseQuery,li2024enhancing,koo2024flexiedit} show that the initial Gaussian noise strongly affects semantic layout, including object count, scale, and spatial arrangement.
INITNO~\cite{guo2024initno} optimizes the initial noise to reduce text–image mismatch.
NoiseRefine~\cite{NoiseRefine} learn residual noise to eliminate the need for classifier-free guidance. 
NPNet~\cite{NPNet} injects semantic information into the initial noise to improve generation quality.
NoiseQuery~\cite{NoiseQuery} builds a noise library that stores attribute‑specific noise patterns.
Li~\etal~\cite{li2024enhancing} leverage initial noise to correct diffusion models' insensitivity to numerical relations.
FreeNoise~\cite{FreeNoise} and FreeInit~\cite{wu2024freeinit} fix the initial noise across frames to enhance temporal consistency in video generation, while FlexiEdit~\cite{koo2024flexiedit} relaxes local noise initialization to enable non‑rigid regional editing.
Building on this insight, we do not steer layout at inference; instead, we locally permute the DDIM initial noise during data construction to create pseudo pairs from single images. This produces appearance variation while preserving structure and scene harmony, enabling effective training for our learning‑based object swapping framework.
\section{Methodology}

Given a source image $I_s$ with its mask $M_s$ localizing the source object $O_s$, our \textit{SourceSwap} aims to replace $O_s$ with the reference object $O_r$ from the reference image $I_r$, where the mask $M_r$ indicates the position of $O_r$, generating a target image $I_t$. 
The target $I_t$ should satisfy three requirements: (i) \emph{object fidelity}: the generated object preserves identity details of $O_r$; (ii) \emph{scene fidelity}: the source background $B_s=I_s(1\!-\!M_s)$ are unchanged and the near‑boundary context is minimally perturbed; (iii) \emph{object–scene harmony}: the swapped object adapts to illumination, viewpoint, and style of $I_s$ and matches the morphology of $O_s$ so that physical interactions in the original scene remain plausible.
The core challenge hinges on representing the ternary relationship $\{B_s, O_s, O_r\}$ among the source background, the source object, and the reference object.

To seamlessly replace the reference object $O_r$ into the source background $B_s$, we leverage the fact that the source object $O_s$ is inherently consistent with $I_s$.
We assume that if $O_r$ is sufficiently aligned in state to $O_s$, it can naturally interact with the source background $B_s$, generating a target image with high reference/scene fidelity and object-scene harmony.
Therefore, we reformulate the relationship among $\{B_s, O_s, O_r\}$ as another problem:
learning the \emph{inter‑object alignment} between two different objects ($O_r$, $O_s$) under the same background $B_s$.
Prior inpainting‑style solutions~\cite{pbe,chen2024anydoor,mimicbrush} mask out $O_s$ during training and inference, forcing the model to infer the object state mainly from $B_s$. This weakens inter‑object correspondence and often causes discontinuities near the mask. In contrast, we keep the source object visible during training (\textit{source‑aware learning}), so the model observes full object–scene relations and learns alignment directly between $O_r$ and $O_s$.
Learning the inter‑object alignment ideally requires pairs that expose differences between \emph{distinct} objects under the same background.
However, collecting such training data is costly, and existing pseudo‑pair strategies mostly depict the \emph{same} object across contexts, as shown in~\cref{fig:moti}, which is helpless for inter-object reference. 
To address this, we introduce a simple \emph{initial‑noise perturbation} mechanism that converts any single image into a pseudo pair by altering object appearance while preserving shape and scene harmony. The details are given in the next subsection.

\begin{figure}
	\centering
	\includegraphics[width=\linewidth]{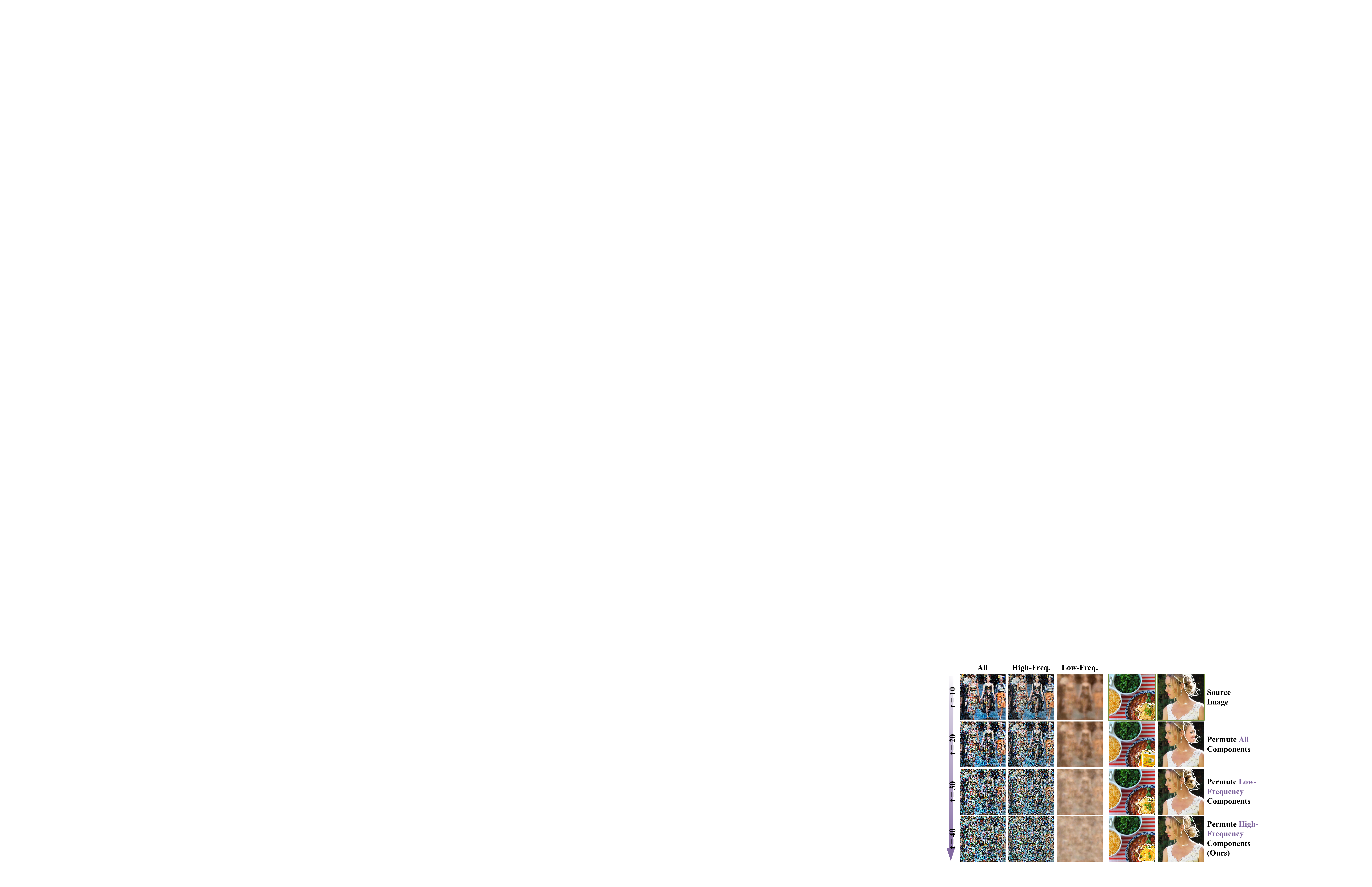}
        % \vspace{-0.3cm}
	\caption{
    Left: decoded DDIM latents at different timesteps and their low and high frequency components. The high-frequency component retains most of the identity and structural cues, while the low-frequency component is coarse.
    Right: ablations during pseudo pair construction. Permuting all components severely distorts the object; permuting only the low-frequency component yields small changes; fixing the low-frequency and permuting only the high-frequency component produces coherent and meaningful appearance edits.
    }
	\label{fig:permute}
    \vspace{-0.6cm}
\end{figure}

\subsection{Initial Noise Perturbation}

\begin{figure}
	\centering
	\includegraphics[width=\linewidth]{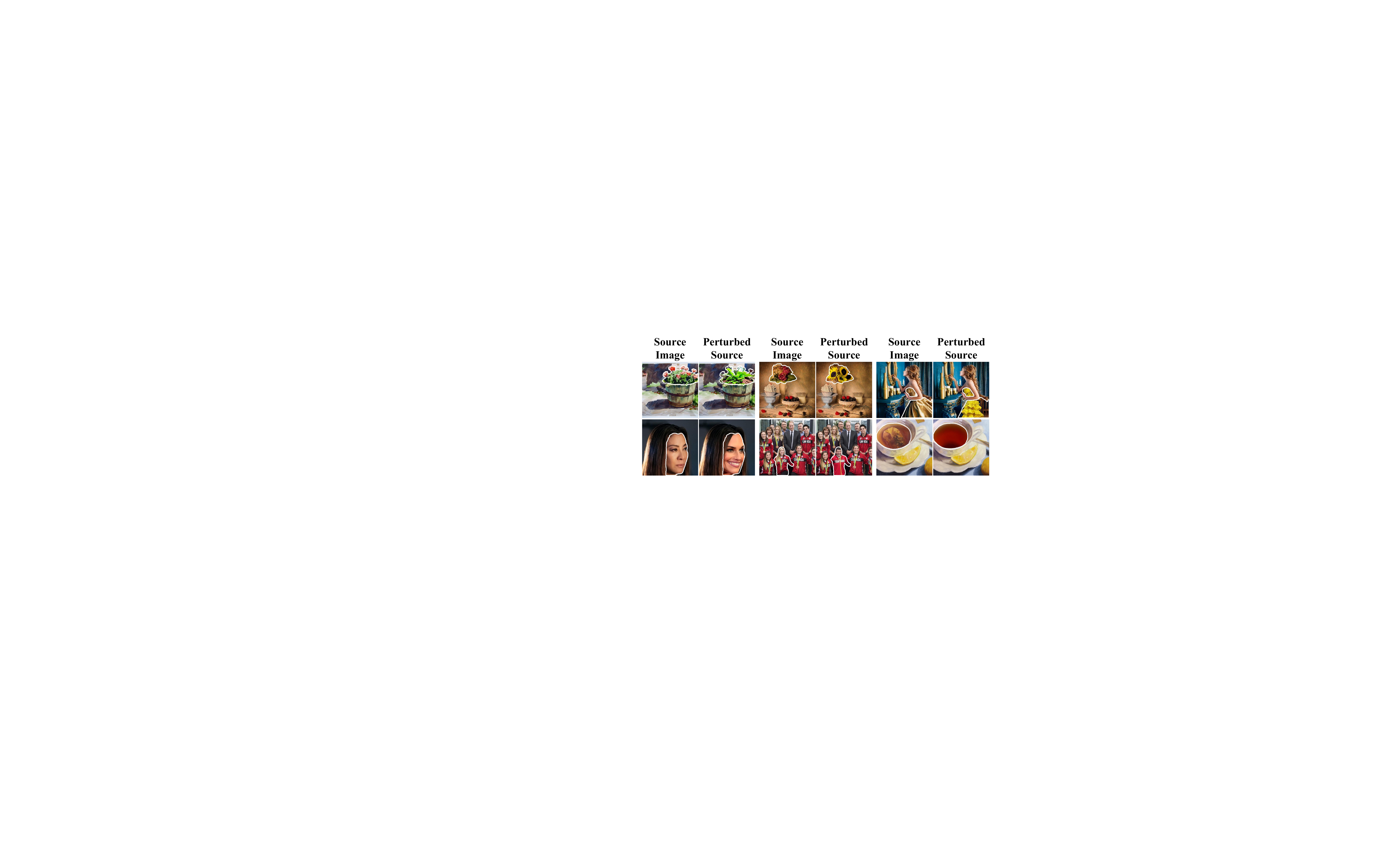}
	\caption{
    Pseudo pairs generated by initial‑noise perturbation. Only high‑frequency latents inside the mask are permuted. The images show appearance variation with preserved structure and scene harmony.
    }
	\label{fig:pair}
    \vspace{-0.3cm}
\end{figure}

To learn the inter‑object alignment without collecting costly inter‑object pairs, we synthesize pseudo pairs from a single image, demonstrated in~\cref{fig:pipeline}. 
The key idea is to alter the \emph{appearance} of the source object while preserving its \emph{coarse shape} and the \emph{scene context}, so that the model can observe a \textit{different object} and learn inter-object alignment with the source object directly.
For a source image $I_s$, we perform DDIM inversion~\cite{ddim} on its latent code $z_0$ to obtain the $T$ step trajectory: \looseness=-1
\begin{equation}
	\{z_t\}_{t=0}^{t=T}=\textrm{DDIM-Inversion}(z_0,\varepsilon_{\theta}, p), \nonumber
\end{equation}
where $\varepsilon_{\theta}$ denotes the denoising U-Net and $p$ is a text embedding that describes the object $O_s$.
We take $z_T$ as the starting point for pair construction.
To alter only the appearance of the object (\eg, material, color, and surface texture) while preserving morphology and layout, 
we separate $z_T$ into low and high frequency parts by a Gaussian low-pass filter:
\begin{equation}
    \begin{aligned}
    z_T^L &= \mathrm{IFFT}(\mathrm{LPF} \odot \mathrm{FFT}(z_T)), \\
    z_T^H &= \mathrm{IFFT}((1-\mathrm{LPF}) \odot \mathrm{FFT}(z_T)), \nonumber
    \end{aligned}
\end{equation}
where FFT and IFFT denote the fast Fourier transform and its inverse, respectively, LPF represents the low-pass filter in the frequency domain, with a stop frequency of 0.3, and $\odot$ indicates the Hadamard product.
Using the source mask $M_s$, we extract the object region $\tilde{z}_T^{H}=M_s\odot z_T^{H}$ and apply a local random permutation over spatial indices while keeping channels aligned:
\begin{equation}
\hat{z}_T^H[c, k] = \Tilde{z}_T^H[c, \pi(k)] \quad \forall c \in \{1,...,C\}, k \in \{1,...,N\}, \nonumber
\end{equation}
where $N = |M_s|$ denotes the number of spatial locations within the mask, and $\pi(\cdot)$ is a random bijective mapping that permutes the spatial indices.
This operation preserves the marginal distribution and energy of the latent while changing local arrangements, which is beneficial for seamless blending. 
In contrast, re‑sampling from a Gaussian prior would mismatch the non‑Gaussian distribution of $z_T$ after inversion and often degrades quality.
Subsequently, $\hat{z}_T^H$ is added to the masked $z_T^L$ in the frequency domain.
In the spatial domain, the background outside the mask is preserved by merging it with the permuted region inside the mask,
yielding the final perturbed initial noise $z_T^P$:
\begin{equation}
    z_T^P =  \mathrm{IFFT}(\mathrm{FFT}(\hat{z}_T^H) + \mathrm{FFT}(M_s \cdot z_T^L)) + (1-M_s) \cdot z_T, \nonumber
\end{equation}
Examples of resulting pairs $(I_s^P, I_s)$ are shown in~\cref{fig:pair}: the object appearance (color, texture, material) changes, while coarse shape and background remain coherent.
\cref{fig:permute} further motivates our design. DDIM latents $z_t$, $z_t^L$, and $z_t^H$ are decoded and shown respectively. The high-frequency component carries the dominant identity and structural cues; permuting all components causes artifacts, and permuting only low-frequency produces changes that are too weak to learn cross-object correspondence.
In addition, permuting all components leads to severe local distortions. Since the low-frequency component is non-dominant, permuting it alone results in minor changes, which are insufficient to learn the inter-object alignment. 
Therefore, we ultimately fix the low-frequency component and permute only the high-frequency component, enabling noticeable yet coherent changes in the object.

It is worth noting that the proposed pseudo-pair construction operates rapidly at low cost. As shown in~\cref{fig:moti}, existing approaches differ significantly: multi-view data acquisition is expensive and often lacks background context; video data is prone to motion blur and requires complex filtering; and instance retrieval depends on costly matching and maintenance of large-scale datasets. While directly cropping an object and background from a single image is straightforward, it fails to capture complex transformations. By contrast, our method produces pairs in which the object changes noticeably while remaining naturally integrated into the same background.
During training, we use $(I_s^P, I_s)$ as a pseudo pair: $I_s^P$ plays the role of the ``source'', and $I_s$ serves as the ``reference''. Combined with source‑aware learning, this exposes direct inter‑object correspondences under a fixed background.

\subsection{Source-Aware Model Structure}
With the constructed paired data, we perform self-supervised training on a dual U-Net architecture~\cite{hu2024animateanyone,ju2024brushnet}, as shown in the right panel of~\cref{fig:pipeline}. 
The upper reference U-Net branch extracts the dense features of the reference object, while the lower denoising U-Net synthesizes the swapping result.
During training, the fine source mask $M_s$ is used only for input preparation.
To increase robustness to imprecise user masks, we first \emph{erode} and then \emph{dilate} $M_s$ 
to obtain a cleaner reference crop, and we expand $M_s$ into a \emph{bounding-box mask} $M_{\mathrm{box}}$ 
to provide coarse spatial localization for the denoiser. 
This design avoids over-constraining the swapped object's shape and improves stability under imperfect masks.
The reference object is then augmented (\eg, blur, elastic transform, perspective transform, zoom in, zoom out) to further simulate various appearances. 
Some methods~\cite{ju2024brushnet,pan2024locate} employing dual U-Net architectures also feed the noisy latent into the reference branch. However, we observe that excluding noise from this branch helps it capture cleaner, more faithful reference details. \looseness=-1

For the lower branch, the noisy source latent is concatenated with the conditional inputs, including the bbox mask $M_{\mathrm{box}}$ and the perturbed source.
Note that the source image serves as the input while the perturbed source acts as the condition, rather than the other way around, as we aim to generate results that approximate the quality of real images, not vice versa.
This \emph{source-aware} design keeps the entire (unmasked) source visible to the denoiser,
so it can:
(i) observe inter-object correspondences jointly with the reference branch,
(ii) preserve object–background interactions from the actual source scene, and
(iii) reduce boundary artifacts caused by imprecise masks.
Finally, each cross-attention block concatenates the Key and Value from the reference network and the denoising network before feeding them into the denoising network.

\begin{figure}
	\centering
	\includegraphics[width=0.9\linewidth]{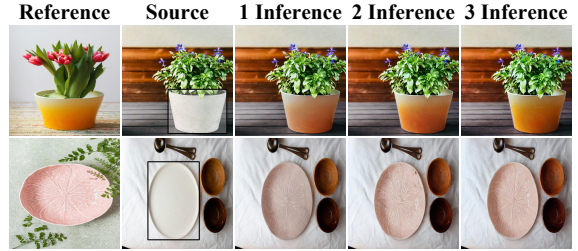}
	\caption{
    Iterative refinement progressively enhances object fidelity by replacing the source image with the output from each inference, recovering fine-grained details such as the vase’s color gradients and the plate’s texture.
    }
	\label{fig:muti_inf}
    \vspace{-0.4cm}
\end{figure}

\begin{figure*}[t]
	\centering
         \vspace{-0.3cm}
	\includegraphics[width=\linewidth]{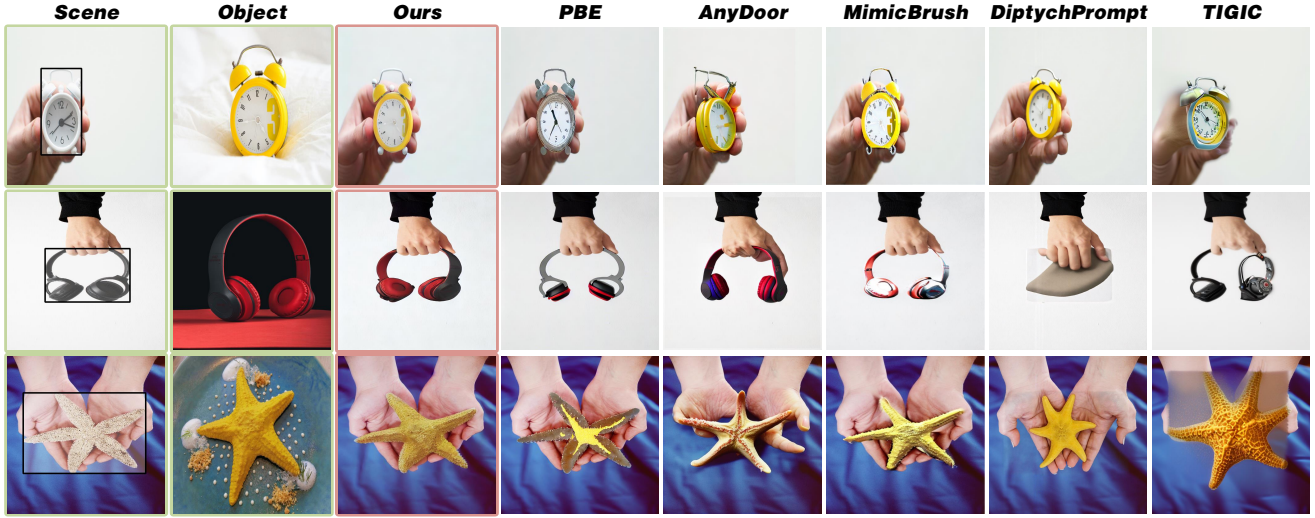}
	\caption{Qualitative comparison with state-of-the-art methods on our SourceBench (please zoom in for a detailed view).}
	\label{fig:comp_sourcebench}
    % \vspace{-0.3cm}
\end{figure*}

\subsection{Train, Inference and Iterative Refinement}
With pseudo pairs $(I_s^P, I_s)$ and the dual U-Net in place, we finetune only the two U-Nets on all constructed pairs without any data filtering.
To make the reference U-Net focus on extracting reference object features, following~\cite{hu2024animateanyone}, we fix its timestep to 0, while the timestep of the denoising network input is uniformly sampled from [0, T].
All other components, including the VAE and text encoder, remain frozen to enable fast, small‑scale finetuning.

At test time, the background of the reference image is removed using Grounded SAM~\cite{ren2024groundedSAM}, and the remaining reference object is fed into the reference branch. 
To further improve object fidelity, we introduce an iterative refinement strategy. 
Since the source-aware inference takes the entire source image as input, the output of the previous iteration can serve as the input for the next:
\begin{align}
    I_s^{(k)} = I_t^{(k-1)}; I_t^{(k)} = \mathcal{D}(I_r, I_s^{(k)}), \nonumber
\end{align}
where $\mathcal{D}$ denotes the source-aware diffusion process, $k$ is the index of iteration.

This strategy boosts the model’s potential without any extra tuning.
\cref{fig:muti_inf} shows that iterative inference progressively enhances object fidelity, especially in color and texture. \looseness=-1

\section{Experiments}

\begin{table}[t]
	\setlength{\abovecaptionskip}{0cm}
	\caption{
    Statistics of SourceBench and DreamEditBench. ``S'' denotes source and ``R'' denotes reference. With a comparable number of pairs, SourceBench contains more object categories/instances and higher resolution than DreamEditBench \cite{lidreamedit}.
    }
	\centering
	\resizebox{0.47\textwidth}{!}{
		\begin{tabular}{l||c|c|c|c|c|c|c}
			\toprule
			\multirow{3}{*}{Benchmark} & \multirow{3}{*}{\begin{tabular}{c}Min.\\Resolution\end{tabular}} & \multirow{3}{*}{\# Class} & \multicolumn{2}{c|}{\# Object} & \multicolumn{2}{c|}{\# Image} & \multirow{3}{*}{\# Pairs} \\
			\cmidrule(lr){4-5} \cmidrule(lr){6-7}
			& & & S & R & S & R & \\
			\midrule	
			\midrule
			DreamEditBench~\cite{lidreamedit} & $168 \times 154$ & 22 & 220& 30 & 220& 158 & 1574\\
			SourceBench & $700 \times 525$ & 54 & 349& 228& 349 & 228 & 1554\\
			\bottomrule
	\end{tabular}}
    \vspace{-0.3cm}
	\label{tab:sourcebench}
\end{table}

\begin{figure*}
	\centering
	\includegraphics[width=\linewidth]{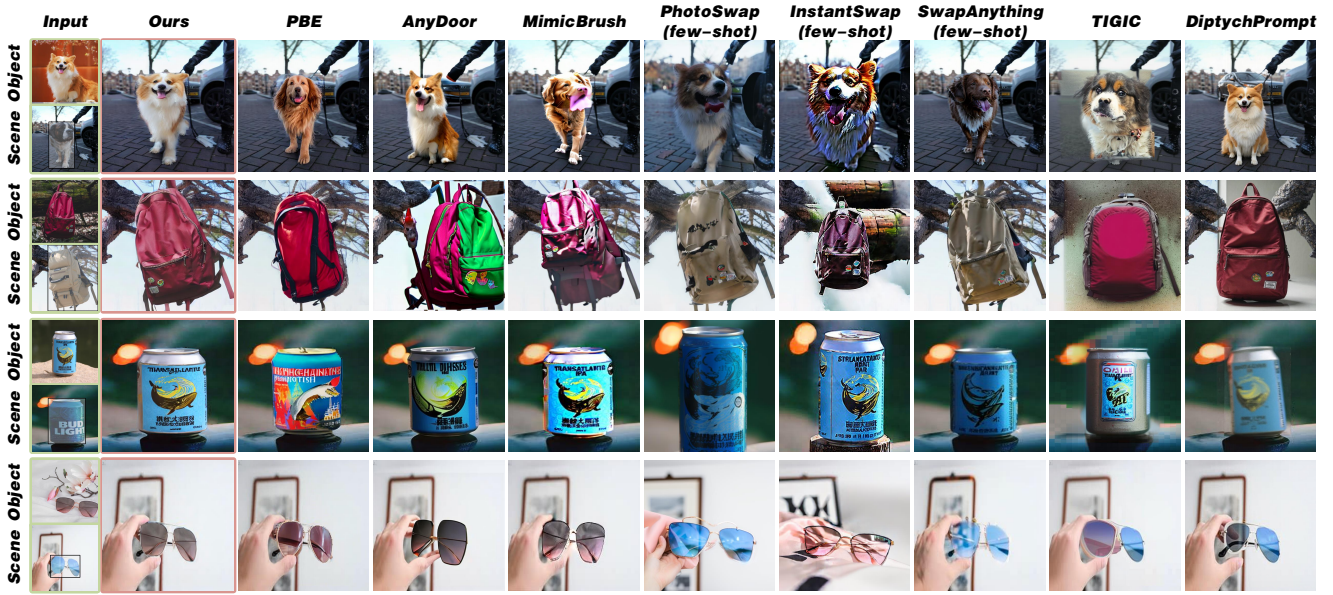}
     \vspace{-0.3cm}
	\caption{Qualitative comparison with state-of-the-art methods on DreamEditBench (please zoom in for a detailed view).}
    \vspace{-0.3cm}
	\label{fig:comp_dreameditbench}
\end{figure*}

\subsection{Settings}
\noindent\textbf{Implementation details.}
During paired data construction, we adopt Stable Diffusion v1.5~\cite{sd} as the base model. Both DDIM inversion and sampling are performed with 50 timesteps,
taking about 52 hours on 4×NVIDIA L40 GPUs in total.
In source-aware training, the reference U‑Net is initialized from Stable Diffusion v1.5 and the denoising U‑Net from Stable Diffusion Inpainting v1.5.
We train for 10000 iterations with a batch size of 4, using the AdamW~\cite{adam} optimizer and a constant learning rate of 1e-5, taking about 8 hours on a single NVIDIA A100 GPU.
During inference, we adopt 20 steps and $k = 2$ rounds of iterative refinement.

% \vspace{-4pt}
\noindent\textbf{Datasets.}
We randomly select $80k$ images from BrushData~\cite{ju2024brushnet}, which is a subset of the LAION‑Aesthetic~\cite{schuhmann2022laion} dataset.
Masks that are excessively large or small relative to their original images are filtered out (\ie, those whose height or width is greater than $\frac{3}{4}$ or smaller than $\frac{1}{5}$ of the corresponding image size).
The remaining $43k$ images are used to generate paired samples, serving as our training data.
% \vspace{-4pt}

\noindent\textbf{Benchmarks.}
For evaluation, since the existing DreamEditBench~\cite{lidreamedit} dataset is built upon DreamBooth~\cite{dreambooth} by simply adding background, it suffers from limited object diversity (more than half categories concentrate on cats, dogs, or toys).
To overcome this limitation, we construct a new high-quality benchmark, \textit{SourceBench}, tailored for object swapping and including complex interactions between objects and backgrounds.
We collect the images from Pexels and Unsplash websites.
With a comparable number of object-scene pairs to DreamEditBench, \textit{SourceBench} provides richer categories and more intricate object-background interactions, enabling more comprehensive evaluation under limited resources (see \cref{tab:sourcebench}). We report results on both SourceBench and DreamEditBench.

% \vspace{-4pt}

\noindent\textbf{Baselines.}
We compare SourceSwap against three categories of existing methods:
(i) learning‑based methods (PBE~\cite{pbe}, AnyDoor~\cite{chen2024anydoor}, MimicBrush~\cite{mimicbrush}),
(ii) test‑time tuning methods (PhotoSwap~\cite{gu2023photoswap}, InstantSwap~\cite{zhuinstantswap}, SwapAnything~\cite{gu2024swapanything}), and
(iii) tuning‑free methods (DiptychPrompt~\cite{diptychprompt}, TIGIC~\cite{TIGIC}).
As test‑time tuning methods cannot perform inference on individual samples and require pre‑training a personalized DreamBooth~\cite{dreambooth} model using few‑shot object images, we report their results only on DreamEditBench.

\subsection{Comparison to Baselines}
\noindent\textbf{Qualitative Evaluation.}
Qualitative evaluation results on two benchmarks are shown in~\cref{fig:comp_dreameditbench,fig:comp_sourcebench}.
PBE roughly aligns the pose between the source and reference objects, but due to the limited representational capacity of CLIP low‑rate embedding, it fails to effectively transfer the appearance of the reference object. AnyDoor enhances feature extraction, but inaccurate masked sources cause the loss of information around object-background boundaries, as seen in the missing tree branches in~\cref{fig:comp_dreameditbench} and the distorted fingers in~\cref{fig:comp_sourcebench}.
MimicBrush tends to produce blurry or over‑exposed results, mainly due to its relatively low quality  training videos.
Tuning‑based methods tend to suffer from underfitting (\eg, PhotoSwap, SwapAnything) or overfitting (\eg, InstantSwap), stemming from the instability of the pre‑trained DreamBooth models. TIGIC, lacking task‑specific training, generates unrealistic results. As another tuning‑free approach, the FLUX‑based~\cite{flux1-dev} DiptychPrompt achieves higher visual quality but still exhibits unnatural transitions around foreground-background boundaries, as shown in the third row of~\cref{fig:comp_sourcebench} and the third row of~\cref{fig:comp_dreameditbench}.
In contrast, our SourceSwap faithfully replaces the reference object at the designated region in the source image while preserving natural interactions between the object and its surroundings (\eg, hands).
This effectiveness benefits from two key factors: the constructed dataset, which facilitates inter‑object correspondence learning, and the source‑aware strategy, which enables the model to directly perceive nuanced object-scene relationships.

\begin{figure}
	\centering
	\includegraphics[width=\linewidth]{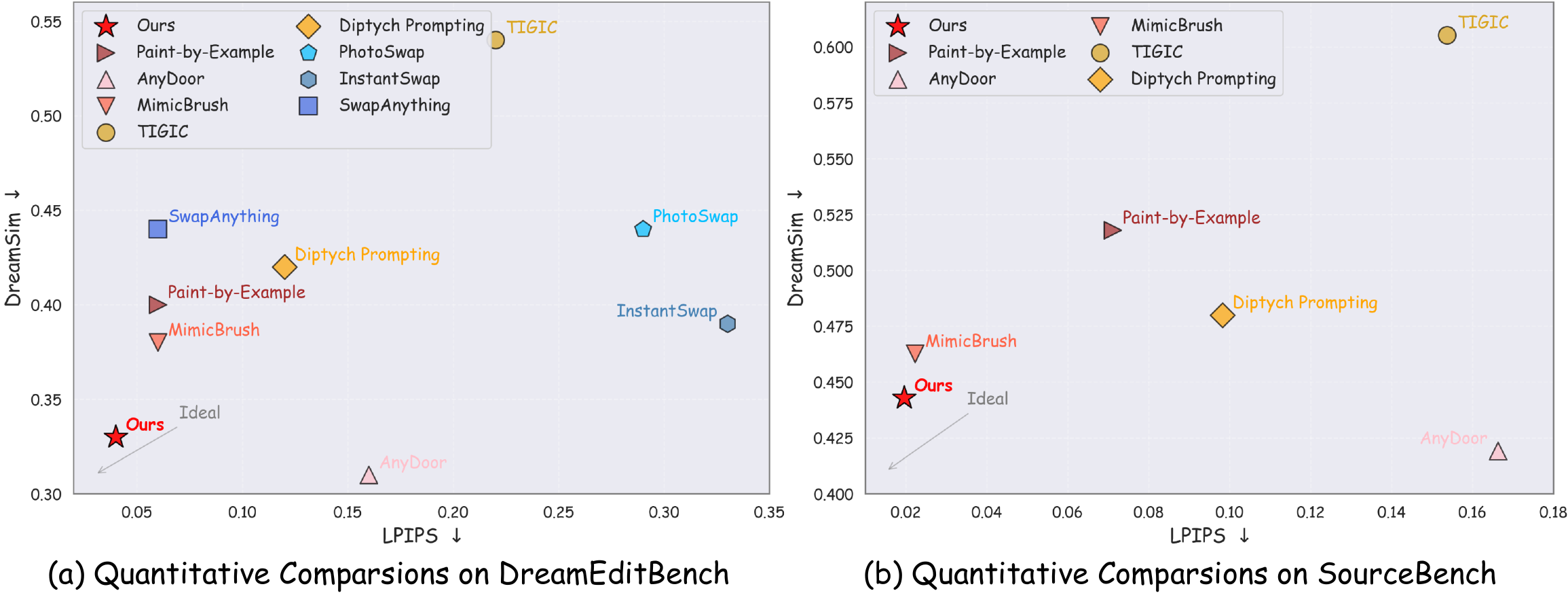}
	% \vspace{-0.4cm} 
    \caption{Quantitative comparisons. Lower DreamSim and LPIPS scores indicate better performance, corresponding to points closer to the lower-left region in (a) and (b).}
	\label{fig:comp_metric}
    % \vspace{-0.4cm}  
\end{figure}
% \vspace{-5pt}
\begin{table}[tb]
	\setlength{\abovecaptionskip}{0cm}
	\caption{
    Evaluation of object-scene harmony using MLLM. Each entry indicates how many times ChatGPT‑5~\cite{gpt5} preferred our method over the competing approaches.
    }
	\centering
	\resizebox{0.4\textwidth}{!}{
		\begin{tabular}{l||c|c}
			\toprule
			{Method}& {DreamEditBench} & {SourceBench}  \\
			\midrule	
			\midrule
			Paint-by-Example~\cite{pbe}& 94.44\%	&  84.31\% \\
			AnyDoor~\cite{chen2024anydoor} 		& 83.33\%   &  68.32\% \\
			MimicBrush~\cite{mimicbrush} 		& 91.67\% 	&  82.35\% \\
			TIGIC~\cite{TIGIC} 			& 97.23\% 	&  96.04\% \\
			Diptych Prompting~\cite{diptychprompt}& 62.86\%	&  65.69\%\\
			PhotoSwap~\cite{gu2023photoswap} 		&  88.89\%  &  - \\
			InstantSwap~\cite{zhuinstantswap} 	&  86.12\%	&  - \\
			SwapAnything~\cite{gu2024swapanything}	&  86.12\%	&  - \\
			\bottomrule
	\end{tabular}}
	\label{tab:comp_mllm}
        % \vspace{-0.3cm}
\end{table}

\noindent\textbf{Quantitative Evaluation.}
We evaluate our method in terms of reference object fidelity, scene fidelity, and object-scene harmony.
To better align with human visual perception, we employ DreamSim~\cite{fu2023dreamsim} to measure the similarity between the reference and generated objects. Compared with DINO~\cite{dino} or CLIP~\cite{clip}, DreamSim captures both high‑level semantics and low‑level visual features, enabling a richer and more fine‑grained visual structures.
For scene fidelity, we use LPIPS~\cite{lpips} as the evaluation metric. Since most background regions can be preserved by directly pasting from the original image~\cite{ju2024brushnet,song2024imprint,mimicbrush}, we focus on the area adjacent to the object. Specifically, we first dilate the fine‑grained mask of the source object and then expand it into a rectangular mask. The complementary region between the rectangular and fine‑grained mask is extracted, and the LPIPS between the source and the generated results are computed within this region.

As shown in~\cref{fig:comp_metric}, our method achieves results closer to the ideal lower‑left region than all baselines and lies on the Pareto front.
Learning‑based methods (MimicBrush, PBE, AnyDoor, and Ours) are generally closer to the Pareto frontier than the other two categories, highlighting that task‑specific data construction effectively benefits subject‑driven generation.
In contrast, tuning‑based methods exhibit a trade‑off between LPIPS and DreamSim, constrained by the inherent limitations of few‑shot tuning.
TIGIC underperforms across both metrics, while DiptychPrompt, despite relying on a stronger base model and carefully engineered prompts, still falls short of the learning‑based approaches.

To evaluate object-scene harmony, we design a Two‑Alternative Forced Choice (2AFC) evaluation based on MLLM. In each trial, two results are provided, one produced by our method and the other by a baseline. 
ChatGPT‑5~\cite{gpt5} first describes both results and then selects the image where the reference object blends more naturally into the scene.
Results are shown in~\cref{tab:comp_mllm}. In all settings, the MLLM consistently prefers our results, indicating that the reference object integrates more naturally into the source scene compared to the competitors.
Moreover, a user study is provided in the supplementary materials.

\begin{figure}
	\centering
	\includegraphics[width=0.8\linewidth]{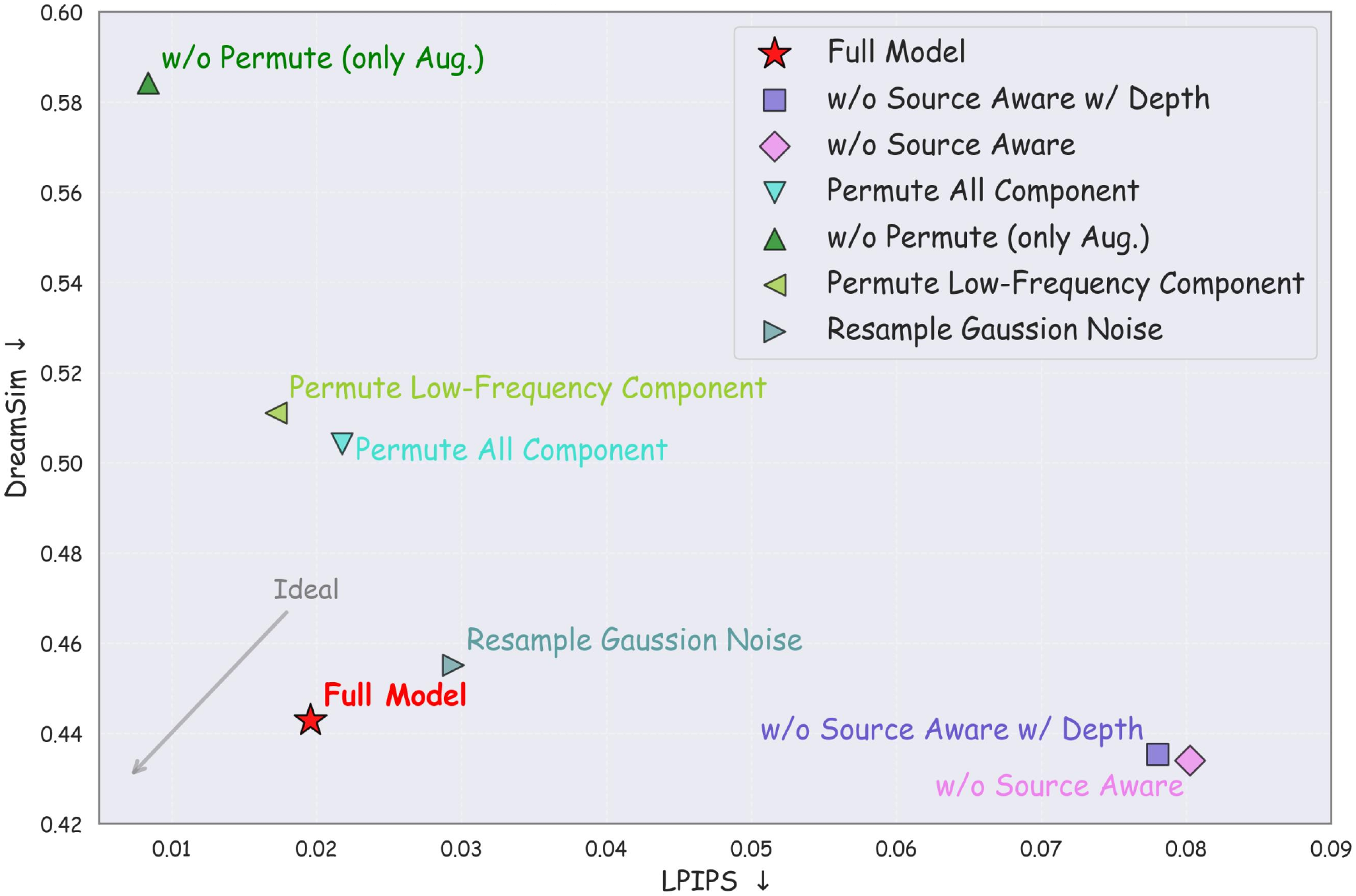}
	\caption{Quantitative ablation study. ``Full model" denotes our complete setting; results closer to the lower-left corner are better.}
	\label{fig:abla_sourcebench_metric}
\end{figure}
% \vspace{-10pt}
\begin{figure}
	\centering
	\includegraphics[width=\linewidth]{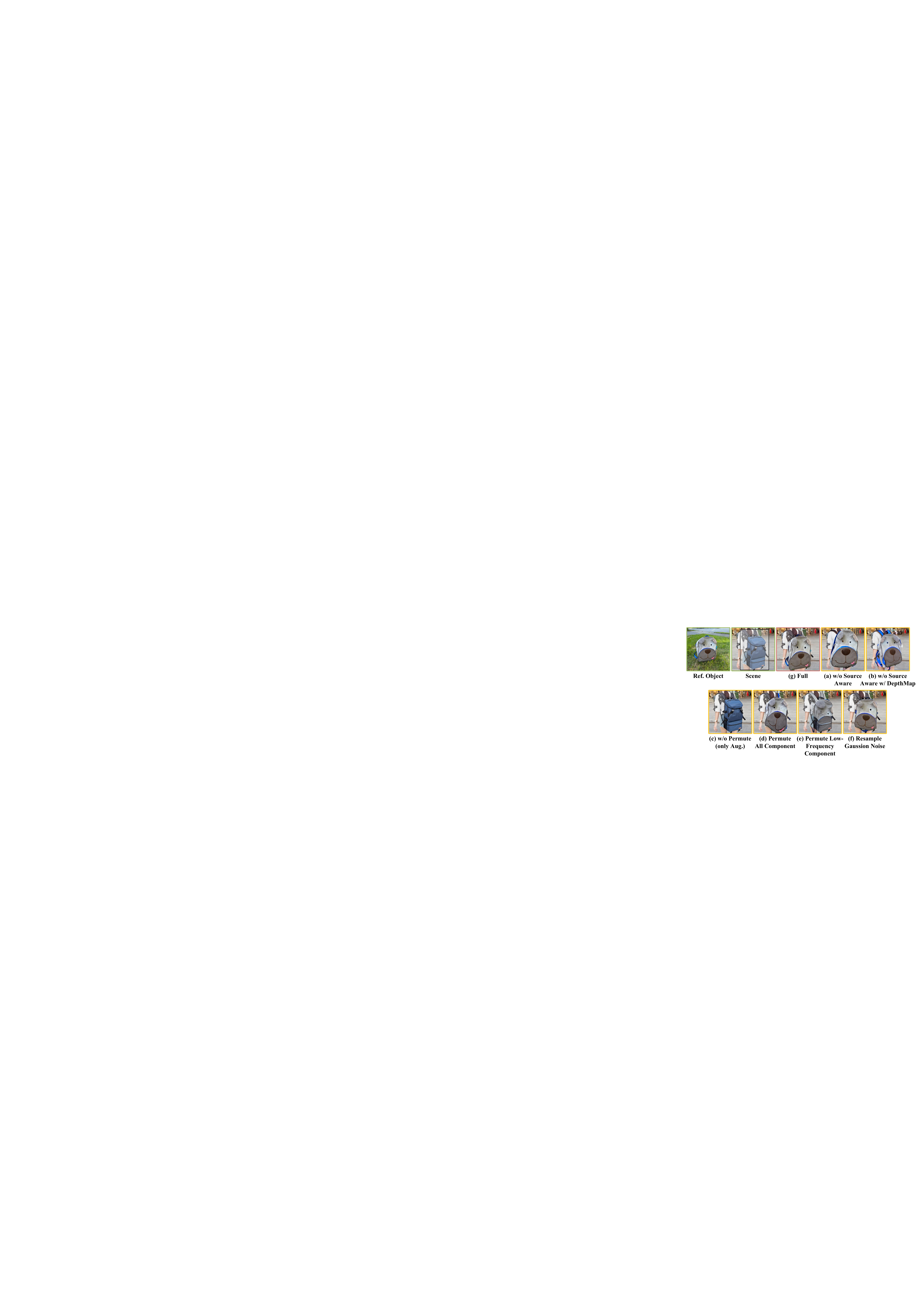}
	\caption{Qualitative ablation study on different variants.}
	\label{fig:abla_fig}
    % \vspace{-0.4cm}
\end{figure}

\subsection{Ablation Study}

Ablation studies are conducted to validate the effect of each component in our framework, including the source-aware strategy and the initial noise perturbation strategy.
We design seven variants: 
(a)~\textbf{w/o Source-Aware}: During training, the perturbed source within the mask is masked out.
(b)~\textbf{w/o Source-Aware w/ DepthMap}: Based on (a), we extract the depth map of the source image using DepthAnything~\cite{yang2024depth} and concatenate it with the input of the denoising U-Net to provide additional guidance.
(c)~\textbf{w/o Permute (only Aug.)}: Trained without paired samples; references are built purely through data augmentation.
(d)~\textbf{Permute All Components}: The frequency-domain perturbation is removed, and all components of the initial noise are locally perturbed in the RGB space.
(e)~\textbf{Permute Low-Frequency Components}: The low-frequency components are permuted instead of the high-frequency ones.
(f)~\textbf{Resample Gaussian Noise}: The initial noise is not perturbed; instead, random Gaussian noise is resampled within the masked regions.
(g)~\textbf{Full Model}: Ours full setting.
The qualitative and quantitative results are shown in~\cref{fig:abla_fig,fig:abla_sourcebench_metric}. 
Moreover, an analysis of the training data size is presented in the supplementary materials.
\looseness=-1

\noindent\textbf{Effect of Source Aware Strategy.}
In \cref{fig:abla_fig} (a), masking the scene image causes the backpack to float incorrectly rather than properly attached to the person. Even when the depth map is introduced as additional guidance in (b), the improvement remains marginal. This incorrect spatial relationship leads to the degradation of LPIPS in \cref{fig:abla_sourcebench_metric}.

\noindent\textbf{Effect of Initial Noise Perturbation.}
We first remove the paired data, as shown in~\cref{fig:abla_fig} (c), which demonstrates that simple data augmentation is insufficient to simulate high-level transform, leading to model collapse.
(d) shows that permuting all components introduces unnatural deformations, since such permutation distorts the structural consistency between data pairs.  
Fixing the low-frequency components effectively prevents this issue.
In (e), although the interaction between the scene and the object is reasonably preserved, the backpack appearance is severely degraded. This is because perturbing only the low-frequency components limits the model’s ability to learn fine-grained details.  
Finally, (f) visually looks like the reference object is simply pasted onto the source image, causing unnatural viewpoint conflicts. This observation aligns with its lower DreamSim and higher LPIPS values in~\cref{fig:abla_sourcebench_metric}.

\section{Conclusion}
We introduce SourceSwap, a framework for high-fidelity object swapping. We propose an initial-noise perturbation that enables single-image pseudo pairing, eliminating reliance on extra data. Our source-aware dual U-Net further learns cross-object alignment directly from full-scene context and supports efficient zero-shot inference with lightweight iterative refinement. We also established SourceBench, a high-quality benchmark with diverse categories and richer interactions. Extensive experiments demonstrate that SourceSwap achieves superior object fidelity, stronger background preservation, and more natural integration, while generalizing well to broader editing tasks such as subject-driven  refinement and face swapping.

\clearpage
% \setcounter{page}{1}
% \maketitlesupplementary
\maketitlesupplementaryWOtitle

\section{Cost Analysis}

\subsection{Training Data Size Analysis}

Benefiting from high-quality pseudo pairs and source-aware training, SourceSwap attains strong performance with only \textbf{40K} images for fine-tuning.
In contrast, prior learning-based baselines require orders of magnitude more data (Table~\ref{tab:trainsize_comp}).
Paint-by-Example~\cite{pbe} uses roughly \textbf{190M} image samples; AnyDoor~\cite{chen2024anydoor} and MimicBrush~\cite{mimicbrush} rely on \textbf{hundreds of thousands} of images plus video frames; IMPRINT~\cite{song2024imprint} and ObjectMate~\cite{winter2025objectmate} further scale to \textbf{millions} of multi-view images, which are expensive to collect and curate.
These data sources not only impose substantial acquisition cost but also depend on modalities beyond single images (videos or multi-view captures).
By comparison, SourceSwap achieves strong performance using only about \textbf{40K} single-image samples.

\begin{figure}[h]
	\centering
	\includegraphics[width=\linewidth]{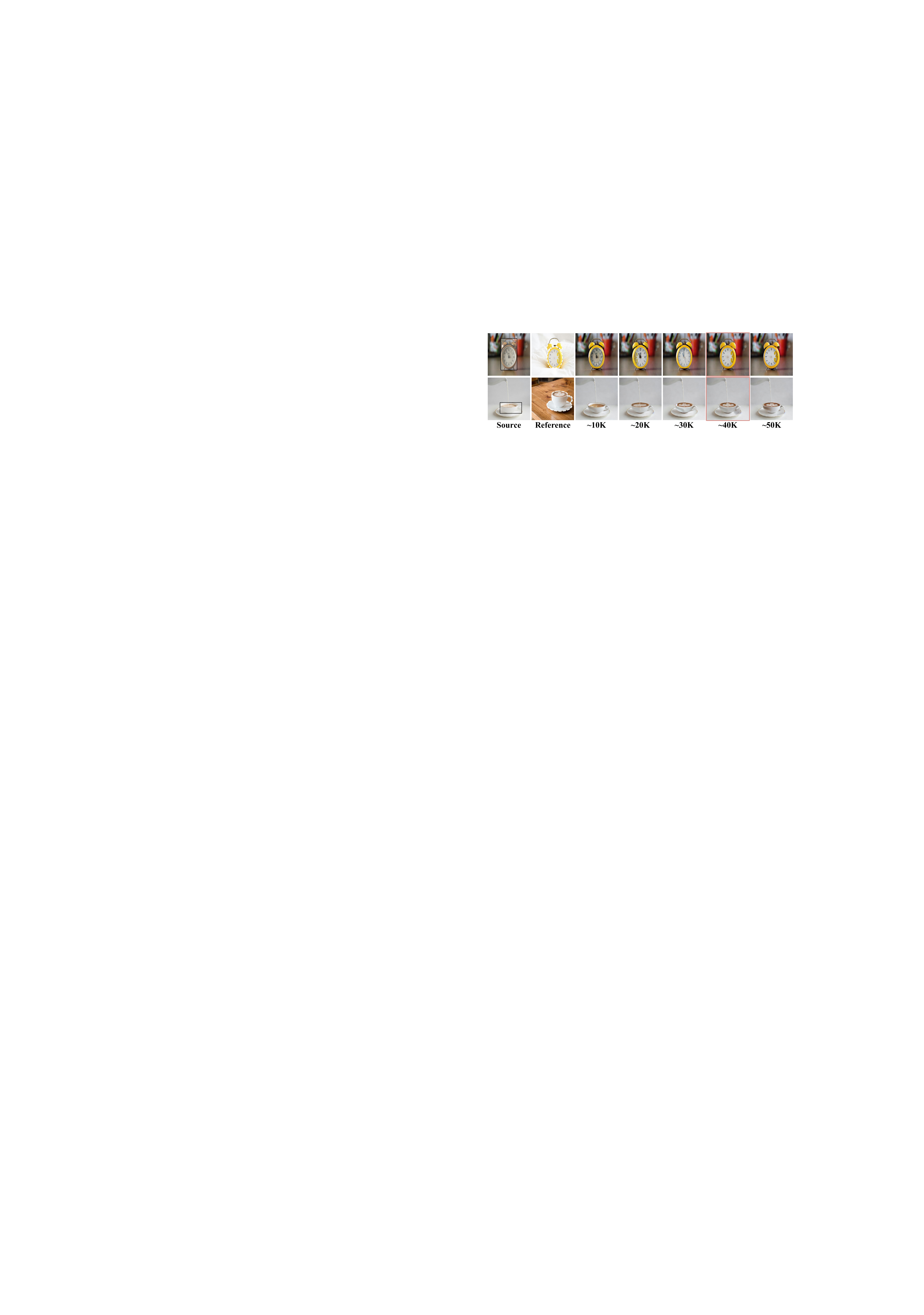}
	\caption{Ablation on training size. Performance saturates at \textbf{40K} samples, indicating that SourceSwap learns effectively from small-scale single-image training}
	\label{fig:abla_trainsize}
\end{figure}

We further conduct an ablation study on the amount of training data, as illustrated in~\cref{fig:abla_trainsize}.
When the training data increases to about 40K, the model already produces high‑quality outputs, while further increasing it to 50K brings only negligible improvement.
Hence, we use about 40K samples for training throughout our experiments.

\begin{table}[h]
	\caption{
    Training data scale of learning-based baselines. Prior methods rely on large and expensive datasets (often videos or multi-view images), whereas SourceSwap requires only 40K single images for fine-tuning.
    }
	\centering
	\resizebox{0.46\textwidth}{!}{
		\begin{tabular}{l||c|c}
			\toprule
			{Baselines} &  {\# Training Data} & {Data Type} \\
			\midrule	
			\midrule
			Paint-by-Example~\cite{pbe} & $\sim\!190000k$ & Images  \\
			AnyDoor~\cite{chen2024anydoor} 		&  $\sim\!410k$ & Images \& Videos \\
			MimicBrush~\cite{mimicbrush} 		&  $\sim\!10100k$ &  Images \& Videos  \\
			ObjectMate~\cite{winter2025objectmate} &  $\sim\!600000k$ &  Multi-view Images  \\
			IMPRINT~\cite{song2024imprint}	& $\sim\!1627k$ &  Multi-view Images \& Videos  \\
			\midrule
			\textbf{Ours} 	&  $\sim\!40k$ & Images \\
			\bottomrule
	\end{tabular}}
	\label{tab:trainsize_comp}
\end{table}

\subsection{Inference Time Analysis}
\label{sec:inference_time}

We report per-sample inference time together with three categories of comparison methods, as shown in \cref{tab:supp_time}. 
All methods are evaluated on a single 40 GB NVIDIA A100 GPU.
Test-time tuning methods take the most time, primarily because a personalized DreamBooth~\cite{dreambooth} model must be trained during inference.
Additionally, PhotoSwap~\cite{gu2023photoswap} and SwapAnything~\cite{gu2024swapanything} depend on precise inversion results, and the 50-step DDIM inversion together with the 500-step null-text optimization~\cite{mokady2023null} introduces significant runtime overhead.
InstantSwap~\cite{zhuinstantswap} uses score distillation to bypass the inversion process, yet its single-image optimization still leads to a higher inference cost compared with learning-based approaches.
For tuning-free methods, TIGIC~\cite{TIGIC} employs a 20-step DDIM inversion, but its three-branch architecture requires three U-Nets to run simultaneously, which limits the overall speed.
DiptychPrompt~\cite{diptychprompt} adopts the FLUX~\cite{flux1-dev} with an additional ControlNet-Inpainting module~\cite{FLUX.1-dev-Controlnet-Inpainting-Beta}, but the 40GB GPU is insufficient to hold all parameters, causing frequent memory swapping between GPU and CPU and thus limiting inference speed.
Learning-based methods require only a forward sampling process, which makes them inherently faster during inference.
Paint-by-Example, AnyDoor, and MimicBrush all use 50 diffusion steps per inference.
In contrast, SourceSwap performs only 20 steps per inference.
\textit{Even with our 2-round iterative refinement, the total cost is merely 4.41 seconds per sample, faster than the most lightweight baseline (Paint-by-Example at 6.33s).}

\begin{table}[h]\centering
\def\arraystretch{1.2}
\small
\scriptsize
\tabcolsep 1.5pt
\resizebox{0.9\linewidth}{!}{
\begin{tabular}{l c c c c c c c c}
  \toprule

    & \multicolumn{8}{c}{\textbf{Test-time Tuning}} \\
    \cmidrule{2-9}
    \multicolumn{1}{c}{} & & \multicolumn{2}{c}{PhotoSwap} & \multicolumn{2}{c}{InstantSwap} & \multicolumn{2}{c}{SwapAnything} &  \\
  \cmidrule{2-9}
  \multicolumn{1}{c}{Time} & & \multicolumn{2}{c}{128.85s(+751.97s)} &  \multicolumn{2}{c}{23.48s(+751.97s)} & \multicolumn{2}{c}{127.39s(+751.97s)} &   \\
  
  \cmidrule{1-9} %\cmidrule{4-7}
\cmidrule{1-9} 
    & \multicolumn{8}{c}{\textbf{Tuning-free}} \\
    \cmidrule{3-9}
    \multicolumn{1}{c}{} & & \multicolumn{3}{c}{TIGIC} & \multicolumn{3}{c}{DiptychPrompt} &  \\
    \cmidrule{2-9}
  \multicolumn{1}{c}{Time} & & \multicolumn{3}{c}{12.84s} & \multicolumn{3}{c}{124.63s}  &   \\

  \cmidrule{1-9} %\cmidrule{4-7}
\cmidrule{1-9} 
    & \multicolumn{8}{c}{\textbf{Learning-based}} \\
    \cmidrule{2-9}
  \multicolumn{1}{c}{}  &  &  \multicolumn{2}{c}{Paint-by-Example} &  \multicolumn{2}{c}{Anydoor} &  \multicolumn{2}{c}{MimicBrush} &  \\
  \cmidrule{2-9} %\cmidrule{4-7}
  \multicolumn{1}{c}{Time}  &  & \multicolumn{2}{c}{6.33s} &  \multicolumn{2}{c}{11.01s} &   \multicolumn{2}{c}{8.47s} & \\ 
  \cmidrule{2-9} %\cmidrule{4-7}
  \multicolumn{1}{c}{} &   \multicolumn{2}{c}{Ours (1 Inf.)} & \multicolumn{2}{c}{Ours (2 Inf.)} & \multicolumn{2}{c}{Ours (3 Inf.)} & \multicolumn{2}{c}{Ours (4 Inf.)}   \\
  \cmidrule{2-9} %\cmidrule{4-7}
  \multicolumn{1}{c}{Time} &  \multicolumn{2}{c}{2.28s} & \multicolumn{2}{c}{4.41s} & \multicolumn{2}{c}{6.64s} &  \multicolumn{2}{c}{8.96s}  \\ 
  
  \bottomrule
\end{tabular}
}
% \vspace{-2mm}
\caption{
Inference time comparison across test-time tuning, tuning-free, and learning-based methods. For test-time tuning methods, we also report (in parentheses) the time required to train the personalized DreamBooth model. For our method, we report 1–4 rounds of inference and adopt 2 rounds as the final setting. 
Learning-based methods are substantially faster, and SourceSwap achieves the best inference speed among them.
}
% \vspace{-2mm}
\label{tab:supp_time}
\end{table}

\section{Analysis of Iterative Refinement}

\begin{figure}[t]
	\centering
	\includegraphics[width=\linewidth]{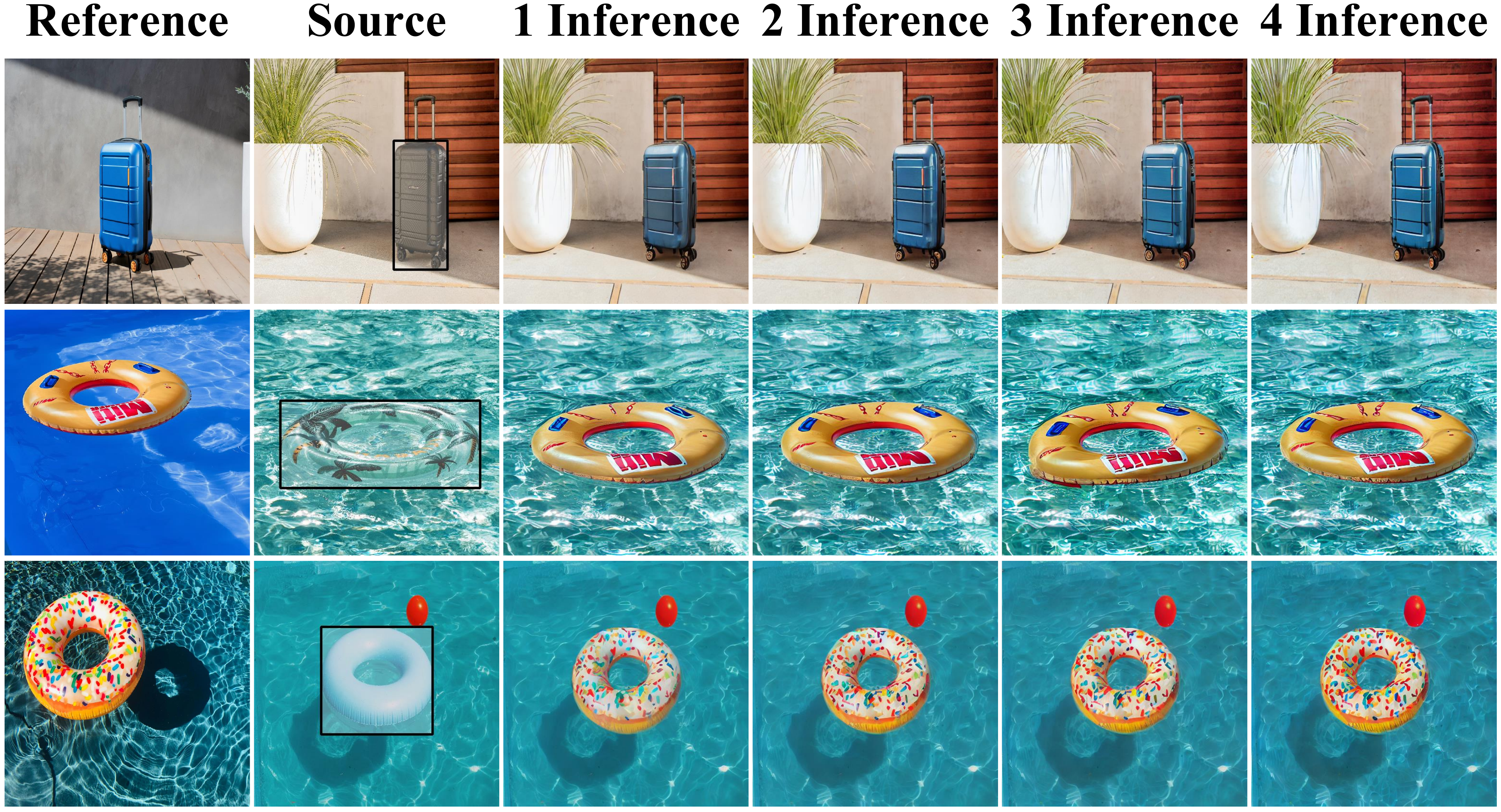}
	\caption{Qualitative effect of iterative refinement. One round ($k=1$) captures the overall structure but misses surface colors; two rounds ($k=2$) significantly improve color consistency, and further rounds ($k=3,4$) yield only marginal visual gains.}
	\label{fig:supp_iterative}
\end{figure}

Our source-aware design enables \textit{iterative refinement}, in which the output of one inference round is fed back as the input for the next.
The influence of the refinement rounds $k$ is summarized in \cref{tab:supp_iterative} and visualized in \cref{fig:supp_iterative}.

As shown in \cref{fig:supp_iterative}, one round of inference captures the overall structure of the reference object but may not fully reproduce its surface color. 
With $k=2$, color consistency and appearance fidelity improve noticeably.
Increasing the rounds to $k=3$ or $k=4$ yields only marginal gains.
\cref{tab:supp_iterative} further confirms this trend: object fidelity generally increases with $k$ and peaks at $k=3$. 
When increasing to $k=4$, DreamSim slightly degrades, which may stem from accumulated compression artifacts introduced by repeated VAE encoding and decoding~\cite{wang2025towards}.
To balance inference efficiency (see \cref{sec:inference_time}) and performance, we report results with $k=2$ in the main paper.

\begin{table}[t]\centering
% \small
% \scriptsize
\resizebox{0.8\linewidth}{!}{
\begin{tabular}{ccccc}
  \toprule
                & {$k=1$} & {$k=2$} & {$k=3$} & {$k=4$} \\
    \midrule
    {DreamSim} & 0.4653 & 0.4428 & 0.4366 &  0.4405  \\
  \bottomrule
\end{tabular}
}
\caption{Effect of iterative refinement rounds $k$ on object fidelity. As $k$ increases, fidelity improves and peaks at $k=3$, while $k=4$ shows a slight drop in DreamSim.}
\label{tab:supp_iterative}
\end{table}

\section{Analysis of Mask Type}

\begin{figure}[h]
	\centering
	\includegraphics[width=\linewidth]{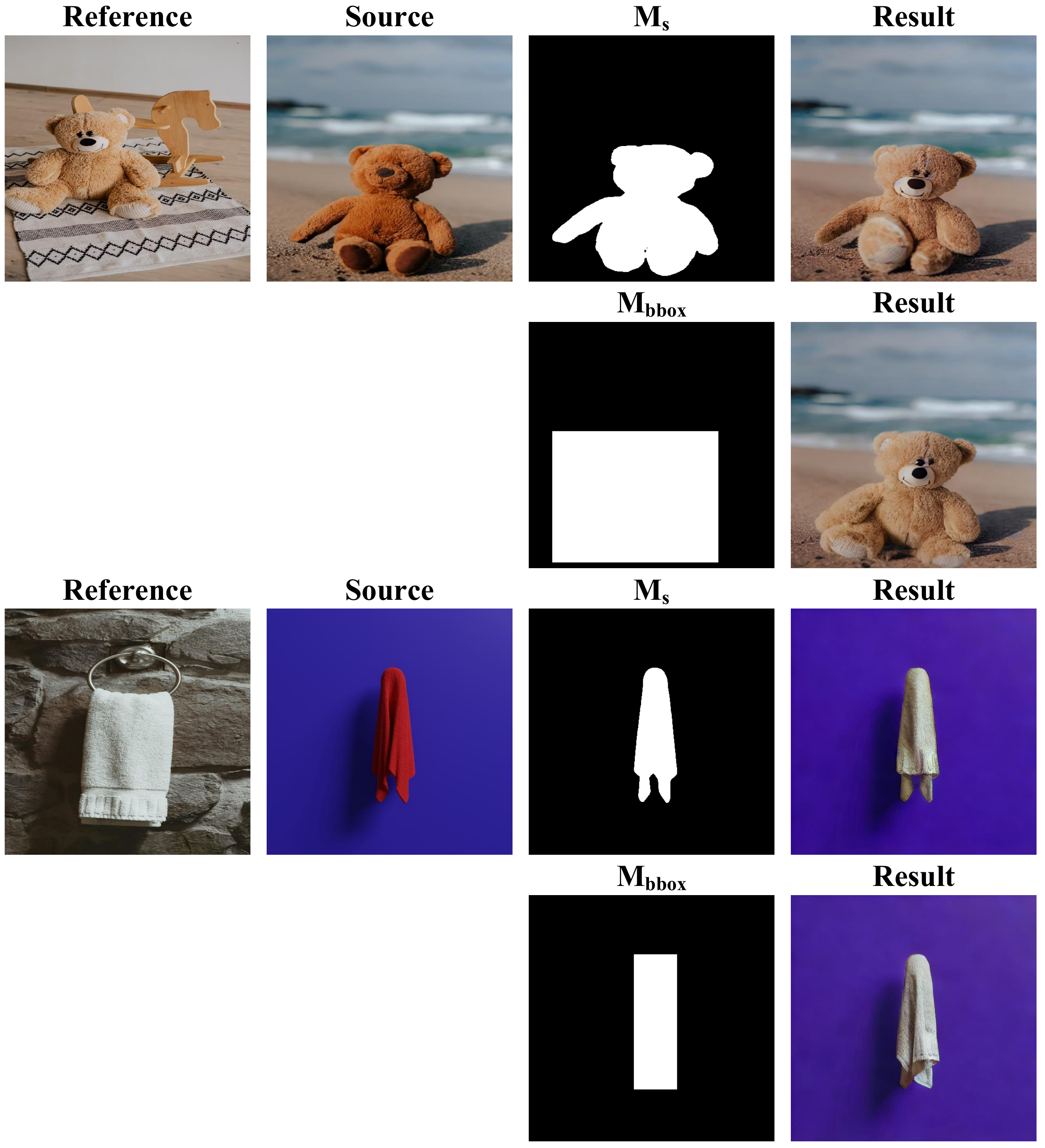}
	\caption{
    Comparison between the precise source mask $M_s$ and the bounding-box mask $M_{box}$.
    We train two variants by conditioning the denoising U-Net on $M_s$ and $M_{box}$, respectively.
    Strictly constraining the object shape with $M_s$ often causes undesirable deformations, whereas $M_{box}$ relaxes the shape constraint and enables the object to retain attributes from the reference.
    }
	\label{fig:precise_mask}
\end{figure}

During source-aware training, the source mask $M_s$ is expanded into a  bounding-box mask $M_{box}$, which is then concatenated to the input of the denoising U-Net as a condition. This expansion relaxes the contour constraints imposed by $M_s$, allowing the resulting object to better preserve attributes of the reference object.
To analyze the impact of mask shape on the results, we train two variants using $M_s$ and $M_{box}$ respectively, as shown in~\cref{fig:precise_mask}. 
Strictly constraining the object shape with $M_s$ can lead to undesirable deformations, for example, thinning of a bear’s arms, distorted leg proportions, and unnatural folding of a towel.
In contrast, using $M_{box}$ provides sufficient spatial guidance while allowing the model to adapt the final object shape according to the reference.
We further quantify this effect using DreamSim: the variant trained with $M_s$ yields 0.4600, worse than the 0.4428 obtained with $M_{box}$, confirming the advantage of relaxed shape constraints.

\section{Limitations}

\begin{figure}[t]
	\centering
	\includegraphics[width=\linewidth]{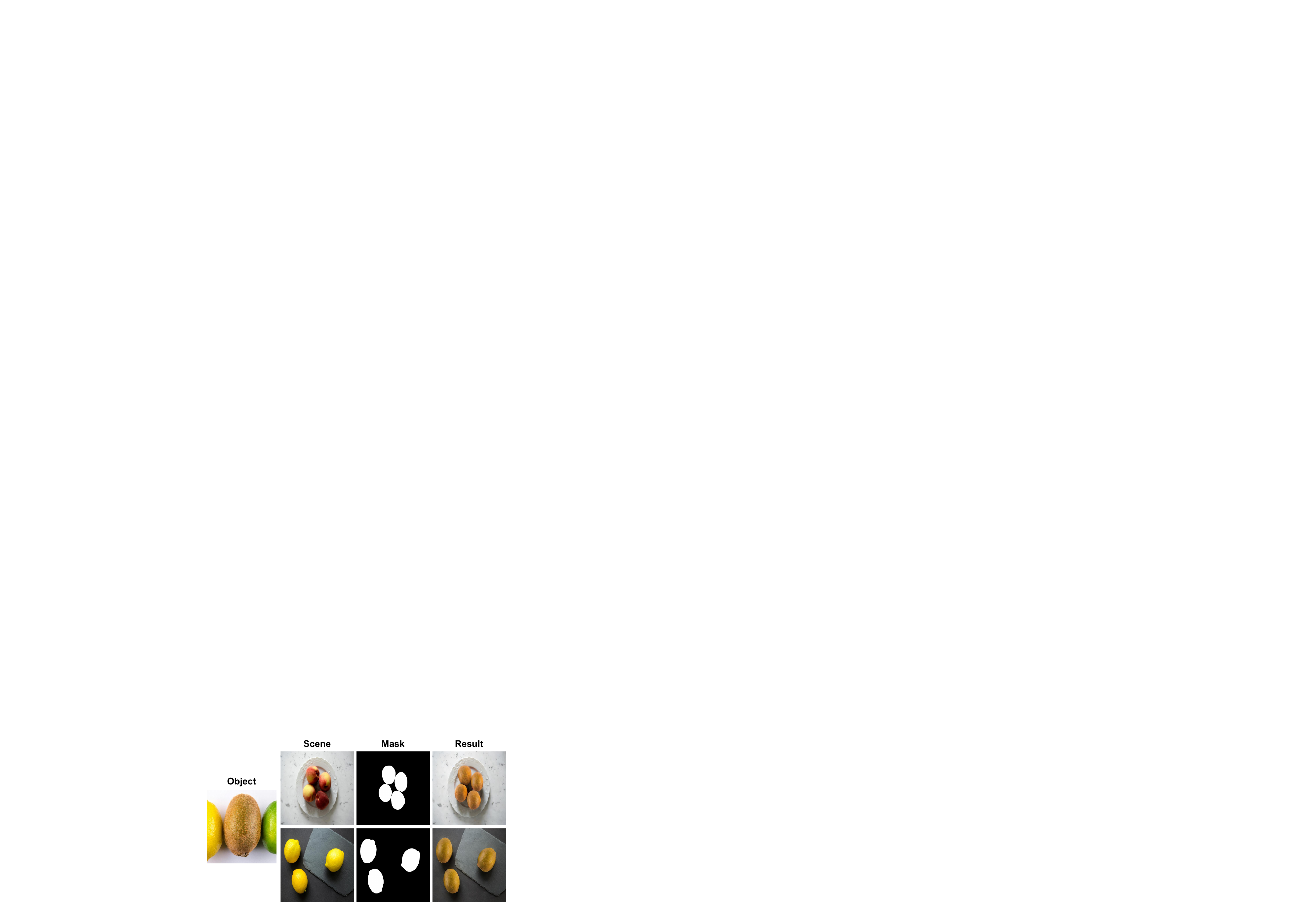}
	\caption{Failure case caused by ambiguous localization. 
Segmentation tools such as Grounded SAM~\cite{ren2024groundedSAM} tend to extract all instances (\eg, multiple lemons), leading the model to swap every segmented object rather than the intended one. 
}
	\label{fig:failure}
\end{figure}

Although SourceSwap performs robustly in diverse scenarios, its accuracy depends on the localization quality of the mask.
Off-the-shelf segmentation tools such as Grounded SAM~\cite{ren2024groundedSAM} tend to segment all objects in the region (\textit{e.g.}, several lemons in \cref{fig:failure}). 
In such cases, the model receives ambiguous control signals and may replace every segmented instance rather than the intended one.
Interestingly, even under this challenging scenario, SourceSwap preserves scene consistency and produces visually coherent swaps, indicating that the learned alignment behavior remains stable despite incorrect masks.
Nevertheless, resolving this limitation requires more precise target specification.
A promising direction is to develop mask-free or instance-aware variants that rely on lightweight point-based interactions (\textit{e.g.}, a single click) to explicitly identify the target object, thereby mitigating ambiguity introduced by multi-instance segmentation.

\section{Applications}

\begin{figure}
	\centering
	\includegraphics[width=\linewidth]{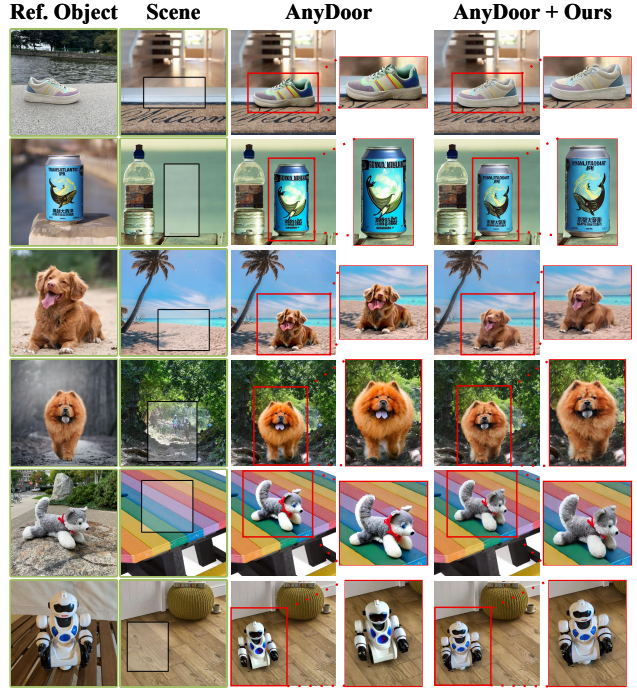}
	\caption{
    SourceSwap is capable of refining suboptimal outputs from other subject-driven generation methods, without additional retraining. 
    It consistently improves the outputs of AnyDoor through its refinement capability.
    }
	\label{fig:app}
\end{figure}

\begin{figure}
	\centering
	\includegraphics[width=\linewidth]{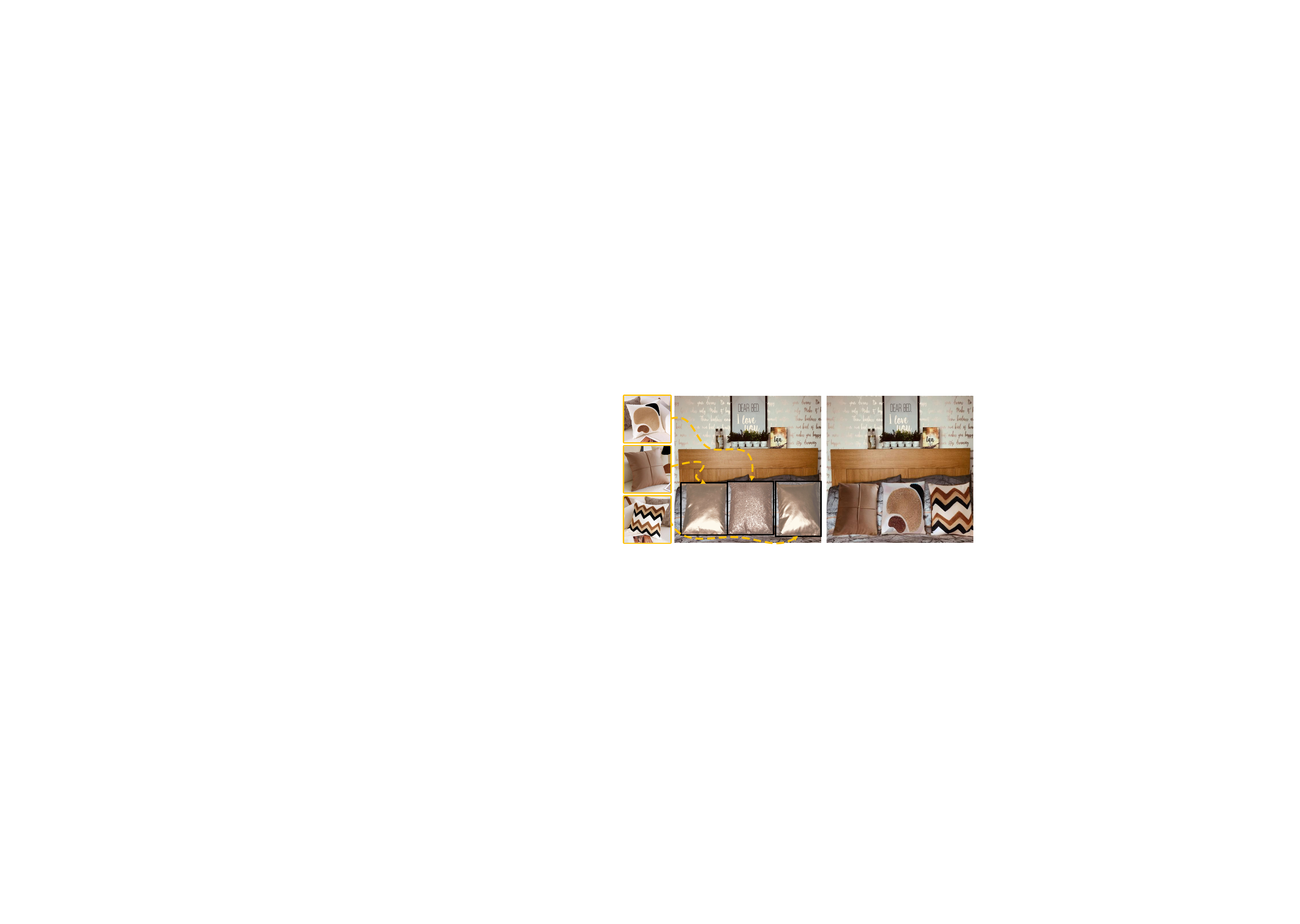}
	\caption{
    SourceSwap is capable of handling multi-object swapping.
    In this example, all three pillows are successfully swapped and blend naturally with the scene.
    }
	\label{fig:muti-subject}
\end{figure}

Our source-aware formulation allows SourceSwap to support a range of applications without any additional retraining, including subject-driven refinement, multi-object swapping, and face swapping.
Given a suboptimal result produced by another method, we simply treat it as the new source image and keep the reference unchanged. The denoising U-Net refines object appearance and restores fine details.
As shown in \cref{fig:app}, our refinement significantly improves the output quality over the original AnyDoor result.
Because each swap operates independently and requires no model-specific tuning, SourceSwap can sequentially replace multiple objects in a single scene.
In~\cref{fig:muti-subject} and the main text, we also present the results of multi-subject swapping. Multiple objects within a single scene are sequentially replaced with three different reference objects. Each object blends harmoniously into the background.
Moreover, by replacing the reference and source with two different faces, SourceSwap is also capable of performing face swapping, as presented in the main text.

\section{User Preference}
\begin{figure*}[h]
	\centering
	\includegraphics[width=\linewidth]{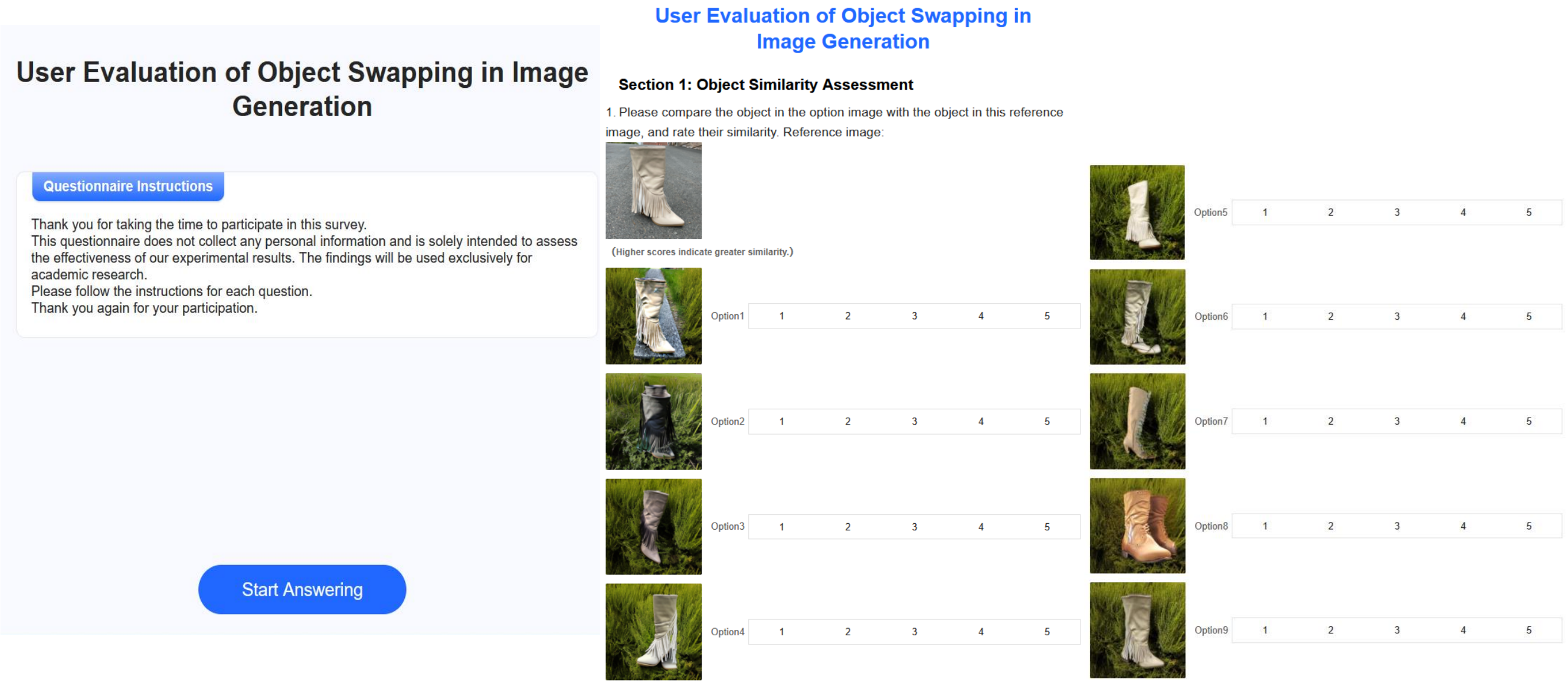}
	\caption{Example of the questionnaire used in the user study. Participants rate each result on object fidelity (object details), scene fidelity (scene preservation), and object–scene harmony using a five-point Likert scale.}
	\label{fig:questionnaire}
\end{figure*}

To evaluate human preference, we conducted a user study using 153 randomly selected results from our method and all competing methods. Participants rated each result on a five-point Likert scale (1–5) across three aspects: \textit{object fidelity} (object details), \textit{scene fidelity} (scene preservation), and \textit{object–scene harmony}.
When object fidelity details, the reference image was shown, and participants were asked to rate how similar the reference object and the result appeared. When rating scene preservation, the source image and result image were presented, and participants were asked to rate how similar the backgrounds were. For object-scene harmony, participants rated how naturally the object blended into the background. 
An example questionnaire is shown in \cref{fig:questionnaire}.
In total, 19 volunteers were recruited, yielding 2,907 answers, as summarized in~\cref{tab:UserStudy}.
Our method achieves the highest scores across all three aspects, reflecting strong alignment with human perception.
Competing methods either underperform across the board or excel only in one or two aspects, indicating limited balance among fidelity, preservation, and harmony.
Notably, the human-judged harmony trend matches the MLLM-based evaluation in Table 2 of the main paper, where our method ranks first, followed by DiptychPrompt~\cite{diptychprompt}.
\begin{table}[htb]
	\caption{
    User study results on three aspects (object fidelity (object details), scene fidelity (background preservation), and object–scene harmony) using a five-point Likert scale (1–5, higher is better).
    }
	\centering
	\resizebox{0.49\textwidth}{!}{
		\begin{tabular}{l||c|c|c}
			\toprule
            \multirow{2}{*}{Baselines} &
            \multicolumn{1}{c|}{Object} &
            \multicolumn{1}{c|}{Background} &
            \multirow{2}{*}{Harmony $\uparrow$} \\
            & Fidelity $\uparrow$ & Fidelity $\uparrow$ & \\
			\midrule	
			\midrule
			Paint-by-Example~\cite{pbe}& 1.63 	& 3.37  & 2.25  \\
			AnyDoor~\cite{chen2024anydoor} 		& 2.74 	& 2.41  & 2.11  \\
			MimicBrush~\cite{mimicbrush}  		& 2.20  & 3.12  & 2.44  \\
			TIGIC~\cite{TIGIC} 			& 1.73 	& 1.95  & 1.53  \\
			Diptych Prompting~\cite{diptychprompt}& 2.62 & 2.68  & 3.15  \\
			PhotoSwap~\cite{gu2023photoswap} 		& 2.26  & 3.23  & 1.96  \\
			InstantSwap~\cite{zhuinstantswap} 	& 2.68  & 1.47  & 1.92  \\
			SwapAnything~\cite{gu2024swapanything}	& 2.39 	& 3.72  & 2.37  \\
			\midrule
			\textbf{Ours} 	& \textbf{3.98} & \textbf{3.82} & \textbf{4.13} \\
			\bottomrule
	\end{tabular}}
	\label{tab:UserStudy}
\end{table}

\section{SourceBench Benchmark Details}
\label{sec:supp_sourcebench}

\subsection{Overview}
SourceBench is a high-quality benchmark designed specifically for object swapping. It complements DreamEditBench by providing higher-resolution, real-world photographs with richer object categories and more complex object–scene interactions. SourceBench contains a comparable number of pairs to DreamEditBench, but significantly more categories and object instances under more challenging conditions. 

\subsection{Data Collection}
Images are collected from Pexels and Unsplash, which provide high-quality, real-world photos under permissive licenses. We discard low-resolution, heavily compressed, or strongly stylized images and retain only images with a minimum resolution of $700 \times 525$ pixels. SourceBench uses single-view images only; no videos or multi-view sequences are involved, so all methods are evaluated purely from still images.

\subsection{Data Filtering Strategy}
\begin{figure*}[h]
	\centering
	\includegraphics[width=\linewidth]{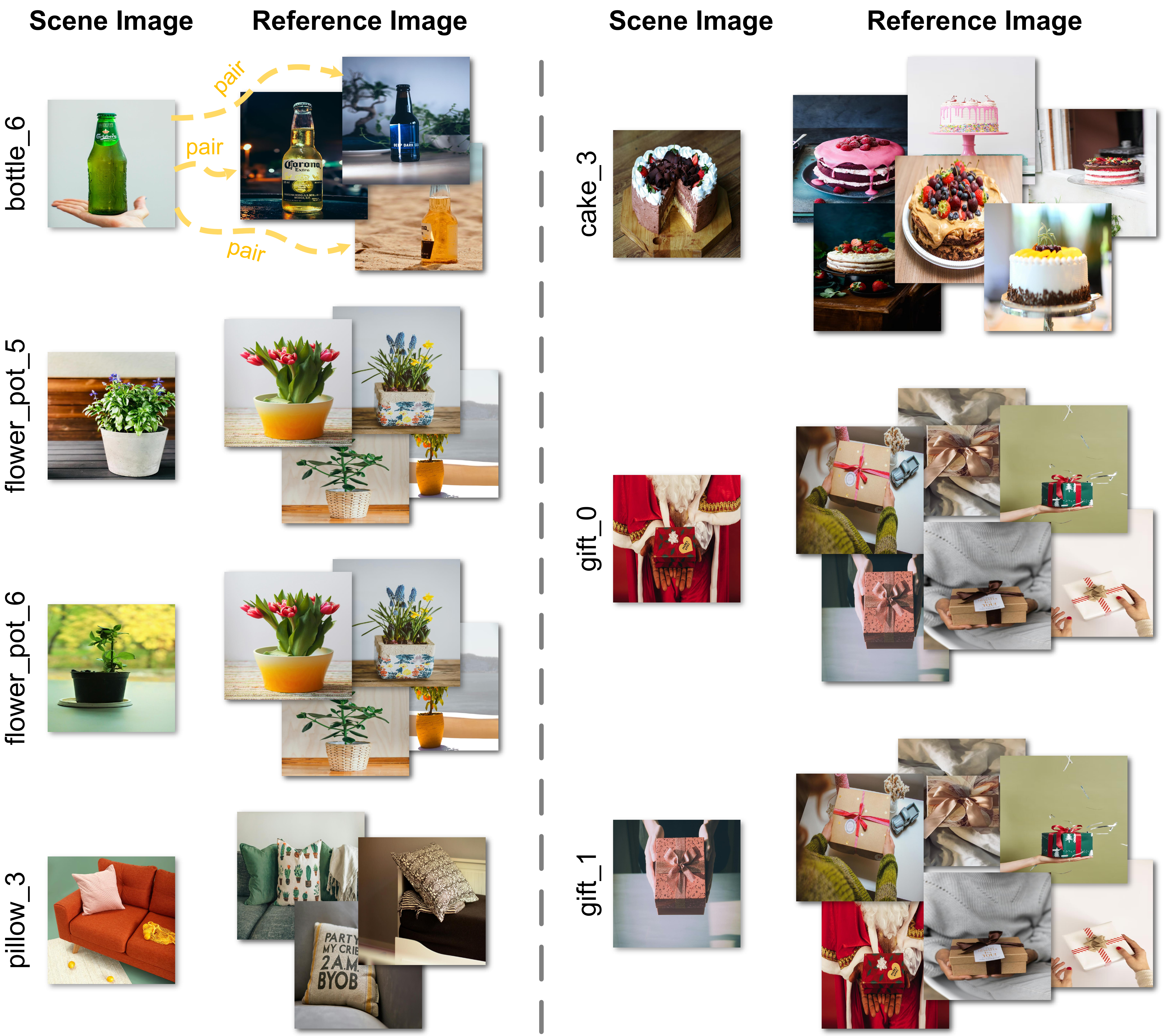}
	\caption{
    Examples of categories and source-reference pairs in SourceBench. 
    Scene images feature rich backgrounds and complex interactions, while reference images provide diverse object appearances. 
    Note that the same image may participate in multiple pairs.
    }
	\label{fig:SourceBench}
\end{figure*}

To make the benchmark informative for evaluating object swapping, we adopt a simple but targeted filtering strategy when selecting background scenes and foreground objects:
\begin{itemize}
    \item For scene images, we avoid images whose backgrounds are overly uniform (\textit{e.g.}, blank walls or nearly textureless backdrops), since such scenes provide little structure for testing object–scene harmony.
    \item For reference images, we avoid choosing objects that are themselves almost textureless or visually trivial, and instead prefer objects with meaningful appearance and geometry (\textit{e.g.}, mugs, books, electronics, tools).
    \item Some images are used in multiple pairs, acting as source in one pair and reference in another, or providing different objects in different pairs. This reuse increases diversity of roles and contexts without inflating the dataset size, as illustrated in \cref{fig:SourceBench}.
\end{itemize}

\subsection{Pair Construction and Statistics}
Each pair in SourceBench consists of a source image, a reference image, and their corresponding masks. The source provides the scene and the object to be replaced, while the reference specifies the target appearance. We manually construct pairs to cover a wide range of everyday categories and interaction patterns, including hand–object interactions, objects placed on furniture, and partially occluded objects. Overall, SourceBench includes 54 categories, 577 distinct object instances, and 1,554 source–reference pairs, offering a more diverse and challenging testbed than DreamEditBench with a similar number of pairs.

We use SourceBench with the same evaluation protocol as in the main paper (DreamSim for object fidelity, local LPIPS for scene preservation, and MLLM/user studies for object–scene harmony), enabling fair comparison across benchmarks.

\section{Multi-modal Large Language Model Evaluation Protocol}
\label{sec:supp_mllm}

To assess the perceptual quality of object-scene harmony beyond conventional metrics, we employ a Two-Alternative Forced Choice (2AFC) evaluation protocol using ChatGPT-5~\cite{gpt5} as the judge. This approach leverages the advanced vision-language understanding capabilities of MLLMs to evaluate nuanced visual qualities such as lighting consistency, viewpoint alignment, and physical plausibility of object-background interactions.

\noindent\textbf{Input Format.}
For each test case, we construct a concatenated image containing four sub-images arranged horizontally:
\begin{enumerate}
    \item \textbf{Source Image} ($I_s$): The source scene with the object to be replaced.
    \item \textbf{Reference Image} ($I_r$): The image containing the reference object to be swapped in.
    \item \textbf{Generated Output 1}: The result produced by our SourceSwap method.
    \item \textbf{Generated Output 2}: The result produced by a baseline method.
\end{enumerate}

The order of outputs (positions 3 and 4) is randomized to mitigate position bias in the MLLM's responses.
We employ a sequential two-stage prompting approach to ensure thorough evaluation:

\noindent\textbf{Stage 1: Description.}
The MLLM is first instructed to describe each of the two generated outputs:

\begin{quote}
\textit{``Imagine you're an expert in image generation. In the set of four images below, the first is the source image and the second is the reference image. Our goal is to replace the object in the source with the object from the reference. The last two images are the generated outputs. First, describe each of the last two images.''}
\end{quote}

This stage serves two purposes: (i) it ensures the model carefully examines both outputs before making a judgment, and (ii) it provides interpretable evidence for the subsequent selection.

\noindent\textbf{Stage 2: Selection.}
Building upon the descriptions from Stage 1, the MLLM is then asked to make a preference judgment:

\begin{quote}
\textit{``Second, choose the better of the two outputs, judging by how naturally the reference object blends into the source image.''}
\end{quote}

The conversation history from Stage 1 is maintained to provide context for this decision. This two-stage approach encourages more deliberate and justified evaluations compared to direct preference queries.

\noindent\textbf{Post-processing.}
To parse the responses from Stage 2, we query once more for each case to extract a definitive choice using the structured model output feature provided by the OpenAI API.
We ask the model to parse the response from Stage 2 into a decision and confidence (0-1) tuple.
The average confidence of all the responses is 0.939.
Note that no image is provided in this parsing step; the decision and confidence are inferred solely from the text response.

For each baseline method, we compute the percentage of test cases where ChatGPT-5 preferred our SourceSwap method over the competitor. The results reported in the main paper represent these preference rates across all evaluated pairs for each benchmark.

\section{More Visual Results}
\begin{figure*}[h]
	\centering
	\includegraphics[width=\linewidth]{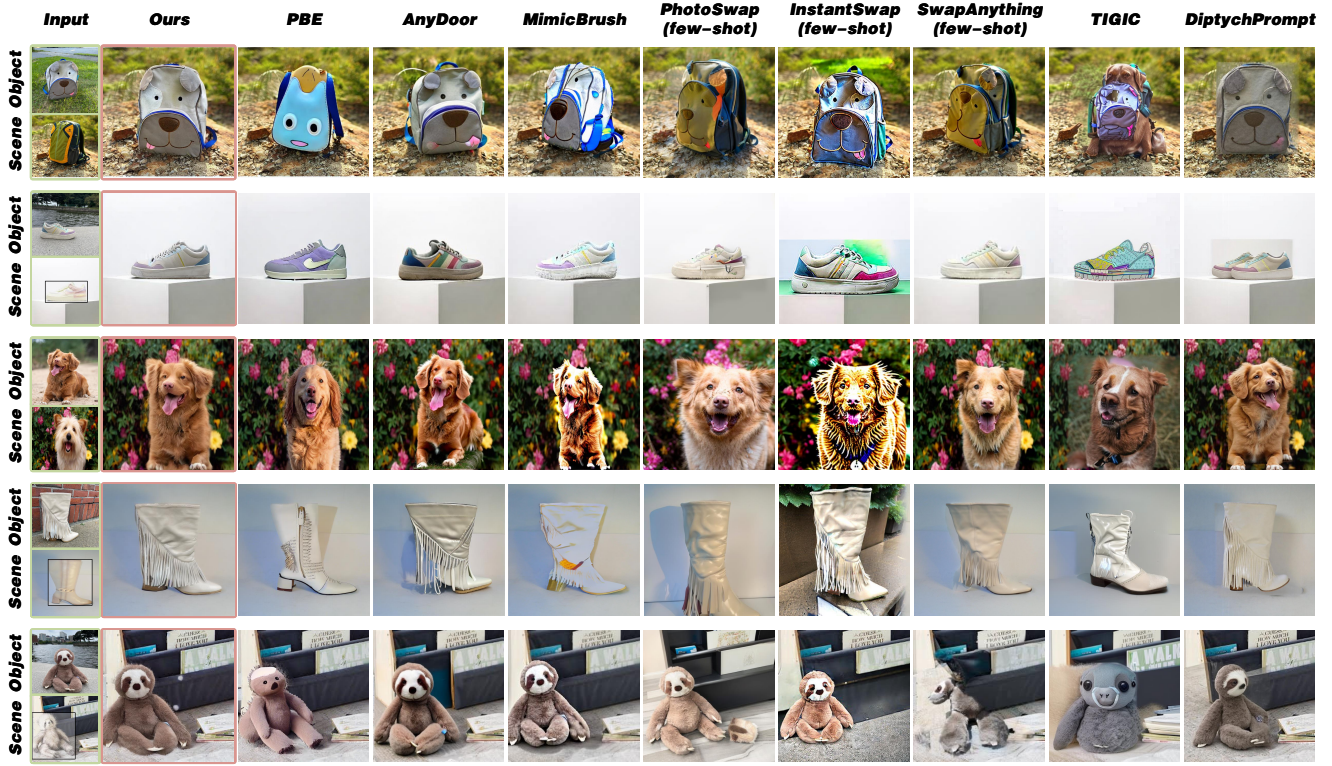}
	\caption{Qualitative comparison with baselines on DreamEditBench.}
	\label{fig:comp_dreameditbench_supp}
\end{figure*}

\begin{figure*}[h]
	\centering
	\includegraphics[width=\linewidth]{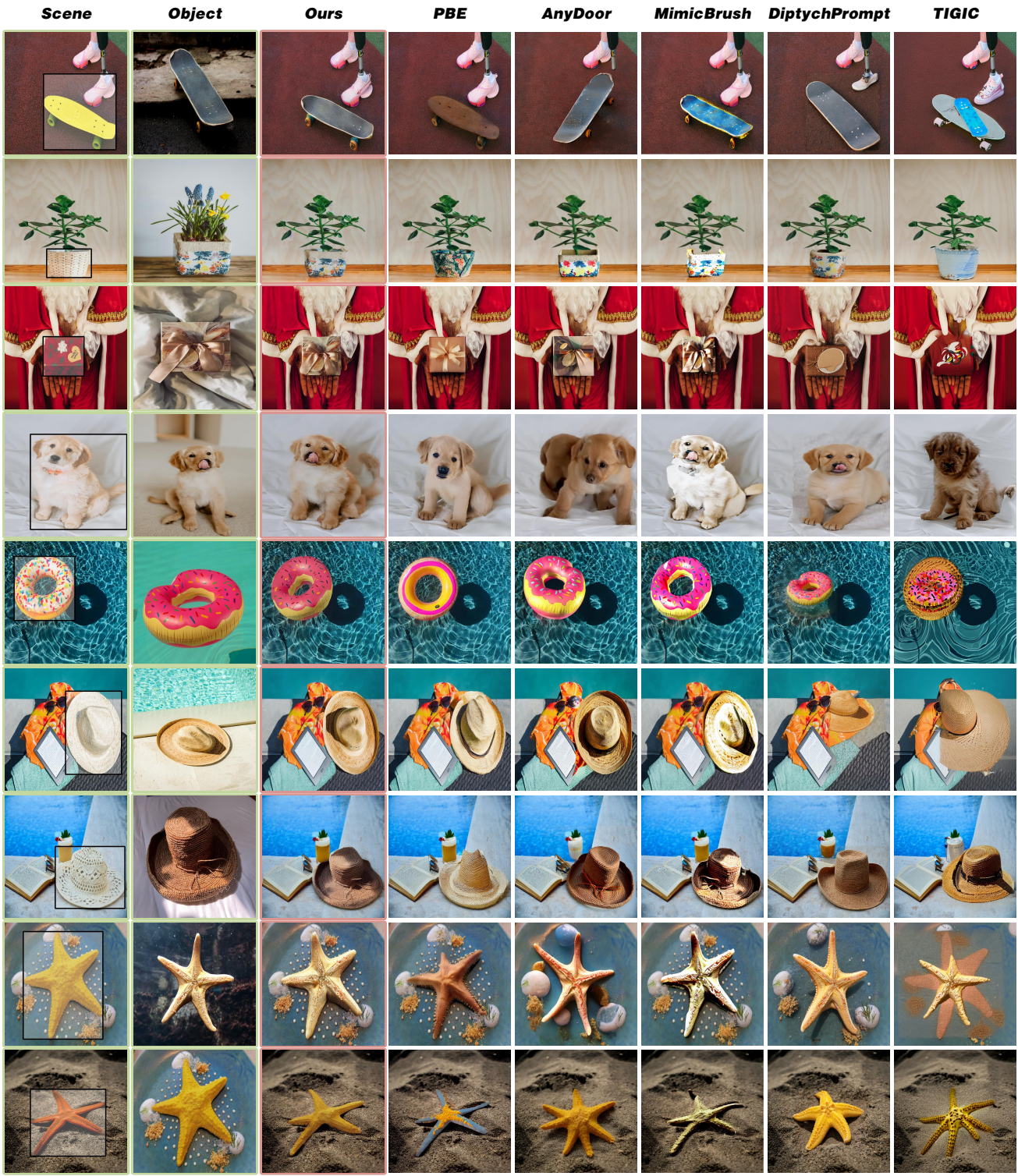}
	\caption{Qualitative comparison with baselines on SourceBench.}
	\label{fig:comp_newbench_supp}
\end{figure*}

\cref{fig:comp_dreameditbench_supp} and~\cref{fig:comp_newbench_supp}  present additional qualitative comparisons between SourceSwap and the baselines on DreamEditBench and SourceBench, respectively. 
The examples cover diverse object shapes, viewpoints, and interaction relationships, demonstrating the robustness of our method under various scenarios.
{
    \small
    \bibliographystyle{ieeenat_fullname}
    \bibliography{main}
}


\end{document}